\documentclass[acmsmall]{acmart}

\usepackage{longtable}
\usepackage{graphicx}
\usepackage{caption}
\usepackage{subcaption}
\usepackage{multirow}
\usepackage{longtable}
\def\BibTeX{{\rm B\kern-.05em{\sc i\kern-.025em b}\kern-.08emT\kern-.1667em\lower.7ex\hbox{E}\kern-.125emX}}

\usepackage{tabularx}
\newcolumntype{L}[1]{>{\raggedright\arraybackslash\textsc{}}p{#1}}

\copyrightyear{2022}
\acmYear{2023}
\setcopyright{acmlicensed}
\acmConference[Woodstock '18]{Woodstock '18: ACM Symposium on Neural Gaze Detection}{June 03--05, 2018}{Woodstock, NY}
\acmBooktitle{Woodstock '18: ACM Symposium on Neural Gaze Detection, June 03--05, 2018, Woodstock, NY}
\acmPrice{15.00}
\acmDOI{}
\acmISBN{978-1-4503-9999-9/18/06}

\begin{document}

%
\title{Numerical Association Rule Mining: A Systematic Literature Review}

%
\author{Minakshi Kaushik}
\email{minakshi.kaushik@taltech.ee}
\orcid{0000-0002-6658-1712}
 \author{Rahul Sharma}
\email{rahul.sharma@taltech.ee}
\affiliation{%
  \institution{Tallinn University of Technology}
  \streetaddress{Akadeemia tee 15a, 12618}
  \city{Tallinn, Estonia}}
 
\author{Iztok Fister Jr.}
\affiliation{%
 \institution{University of Maribor}
 \streetaddress{Koroska cesta 46, SI-2000}
 \city{Maribor, Slovenia}
}
\email{iztok.fister1@um.si}
\author{Dirk Draheim}
\email{dirk.draheim@taltech.ee}
\affiliation{%
  \institution{Tallinn University of Technology}
  \streetaddress{Akadeemia tee 15a, 12618}
  \city{Tallinn, Estonia}}

%
\renewcommand{\shortauthors}{Minakshi et al.}



\begin{abstract}
Numerical association rule mining (NARM) is a widely used variant of the association rule mining (ARM) technique, and it has been extensively used in discovering patterns in numerical data. Initially, researchers and scientists incorporated numerical attributes in ARM using various discretization approaches; however, over time, a plethora of alternative methods have emerged in this field. Unfortunately, the increase of alternative methods has resulted into a significant knowledge gap in understanding diverse techniques employed in NARM -- this paper attempts to bridge this knowledge gap by conducting a comprehensive systematic literature review (SLR). We provide an in-depth study of diverse methods, algorithms, metrics, and datasets derived from 1,140 scholarly articles published from the inception of NARM in the year 1996 to 2022. Out of them, 68 articles are extensively reviewed in accordance with inclusion, exclusion, and quality criteria. To the best of our knowledge, this SLR is the first of its kind to provide an exhaustive analysis of the current literature and previous surveys on NARM. The paper discusses important research issues, the current status, and the future possibilities of NARM. On the basis of this SLR, the article also presents a novel discretization measure that contributes by providing a partitioning of numerical data that meets well human perception of partitions.

\end{abstract}

%
%


%

%

%
\maketitle

\section{Introduction}

Decision-makers have used a wide variety of data mining techniques to extract valuable insights from data. Out of these techniques, association rule mining (ARM) is one of the established data mining techniques. ARM was first proposed by R. Agrawal \cite{Agrawal1993}, and it is primarily used to identify interesting relationships between various data items, e.g., market basket analysis. Later, it has also been used in medical diagnosis and bioinformatics.


In the original settings of ARM, classical algorithms such as Apriori \cite{agrawal1994fast}, Eclat \cite{zaki2000scalable} and FP-growth \cite{han2004mining} were limited to work with boolean datasets only and do not support numerical data items like height, weight, or age. To extend the scope of ARM to support numerical items, Srinkant et al. \cite{srikant1996} proposed a new technique ``quantitative association rule mining (QARM)''. In this technique, numerical data items are converted to categorical data through a discretization process. In literature, QARM  is also referred to as ``numerical association rule mining (NARM)''  \cite{altay2020intelligent}. 


In the early stages of research on NARM, researchers and scientists have used various discretization approaches. However, as time progressed, a wide range of alternative methods emerged, offering novel and innovative solutions in this field. Unfortunately, the increased number of alternative methods has created a substantial knowledge gap, making it difficult to fully comprehend the diverse range of techniques utilised in NARM. 

To address this knowledge gap, this paper conducts a comprehensive systematic literature review (SLR) by following one of the established research methodologies for SLR as outlined by Kitchenham and Charters’s \cite{article}. 
Before conducting this SLR, we thoroughly reviewed several surveys and reviews on NARM, which are listed in Table \ref{reviews}. However, it is important to note that these existing surveys and reviews have certain limitations. They often lack well-defined research questions, comprehensive search strategies, and rigorous research methodologies. Notably, to the best of our knowledge, no SLR of the existing literature on NARM has been conducted to date. The absence of a systematic review in the field has highlighted the need for this article and inspired us to fill this knowledge gap. Indeed, the identified limitations in previous surveys and reviews raised the importance of conducting SLR on NARM. Through this SLR, we aim to address these limitations and fulfil the need for a more comprehensive  understanding of the field. 

In order to provide a complete overview of the NARM literature, we conducted a systematic search across various academic databases and digital libraries to identify relevant scholarly articles. We majorly focused on articles published from the inception of NARM in 1996 up until 2022. In total, we identified 1,140 articles that met our search queries. Next, as per the research methodology, we applied a rigorous process of inclusion, exclusion, and quality assessment criteria to ensure that the selected articles were relevant to the research domain and of high quality. After the screening process, we narrowed down the initial list to a final selection of 68 articles. By following this systematic approach, we aimed to gather a comprehensive and reliable set of articles that contributes to the thorough analysis and synthesis of existing knowledge on NARM.

Based on the exhaustive analysis of 68 articles, this SLR provides an in-depth examination of diverse methods, algorithms, metrics, and datasets utilised in NARM. We thoroughly evaluate the strengths and weaknesses of these methods, algorithms, and metrics while also highlighting their outcomes and potential applications. By conducting a comprehensive analysis of the available literature, we aim to provide deep insights and understanding that can benefit researchers, practitioners, and stakeholders in the field.

As per the findings of this SLR, the article also contributes by introducing an automated novel discretization measure that addresses the human perception of partitions, providing a meaningful and accurate partitioning of numerical data. This novel measure aims to overcome the limitations of existing methods by providing a more meaningful and accurate partitioning of numerical data.

The primary contributions of this paper are as follows.
\begin{itemize}
    \item Well-defined research questions and a methodology for extracting data for a systematic investigation in the area of mining numerical association rules.
    \item Detailed knowledge about NARM methods and their algorithms.
    \item Identified popular metrics to evaluate NARM algorithms.
    \item Identified the major challenges involved in generating numerical association rules, along with some probable future perspectives.  
    \item A novel automated measure is presented for discretizing numerical attributes to contribute to NARM by providing a partitioning of numerical data that meets well human perception of partitions.
    \item Fills the gaps and overcomes the limitations of previous surveys.      
\end{itemize}

The article is organized as follows: Section \ref{BG} presents an overview of the background and related work. In Section \ref{methodology}, we detail our research methodology and articulate the research questions (RQs). Section \ref{result} presents the findings of the review. In Section \ref{threats}, we address the potential threats to the validity of this research article. Section \ref{discussion} delves into a comprehensive discussion of the SLR's findings. Finally, we draw our conclusions in Section \ref{conclusion}.

\begin{table}[!ht]
    \centering
    \caption{Contributions and Limitations of Previous Reviews (N.M.= Not Mentioned)} 
    \label{reviews}
    \begin{tabular} {p{1.7cm}  p{9mm} p{7mm} p{2cm} p{37mm} p{2.8cm}}
    \hline
Paper & Type & Time-frame & Methodology & Contributions & Limitations \\\hline
This Article & SLR & 1996-2022 & Kitchenham's guideline \cite{article} & Present detailed and systematic review encompassing various aspects of NARM, including methods, algorithms, and other relevant factors. & \\
Kaushik et al. (2021) \cite{kaushik2021systematic} & Review & 1996-2020 & Undefined & Investigated discretization techniques in various NARM methods and assessed 30 NARM algorithms. & Authors focused only on important algorithms for three methods.\\

Adhikary et al. (2015) \cite{adhikary2015trends} & Survey & N.M. & Undefined & Authors presented clustering, partitioning, and fuzzy approaches, including evolutionary, statistical and info-theoretic approaches. & This study did not present a detailed study of the discretization method.\\

Adhikary et al. (2015) \cite{adhikary2015mining} & Review & N.M. & Undefined & Discussed applications of NARM. & Approaches were not discussed in detail.\\

Gosain et al. (2013) \cite{gosain2013comprehensive} & Survey & N.M & Undefined & Presented a comparative study of different approaches of ARM including Quantitative data. & Approaches were not categorized and randomly presented.\\ 
\hline
\end{tabular}

\end{table}

\section{BACKGROUND AND RELATED WORK}
\label{BG}
In this section, we provide an in-depth explanation of the background of ARM and NARM.

\subsection{Association Rule Mining}
In the original setting, association rules are extracted from transactional datasets composed of a set $I = \{i_1,\dots,i_n\}$ of $n$ binary attributes called \emph{items} and a set $D = \{t_1,\dots,t_n\}$, $t_k\subseteq I$, of \emph{transactions} called database. An \emph{association rule} is a pair of itemsets $(X, Y)$, often denoted by an implication of the form $X\Rightarrow Y$, where $X$ is the antecedent (or premise), $Y$ is the consequent (or conclusion) and $X\cap Y = \emptyset$. In ARM, support and confidence measures are widely utilized and considered fundamental metrics.
The support of an itemset $X$ determines how frequently the itemset appears in a transactional database. 
The support of an association rule $X \Rightarrow Y $ can be defined as the percentage of transactions among the total records that contain both itemsets \emph{X} and \emph{Y}, shown in Eq. \ref{eq1}.

The confidence of an association rule $X \Rightarrow Y $ determines how frequently items in $Y$ appear in transactions that contain $X$.
The confidence of a rule is calculated as the percentage of transactions that contain itemset \emph{X} also contain itemset \emph{Y}, to the total number of records that
contain $X$ shown in Eq. \ref{eq2}.

\begin{equation}
    \label{eq1}
   Support ( X \Rightarrow Y) = \frac{\lvert(X\cup Y)\rvert}{\lvert D \rvert}
\end{equation}

\begin{equation}
\label{eq2}
 \textit{Confidence}(X \Rightarrow Y) = \frac{\lvert(X\cup Y)\rvert}{\lvert X \rvert}
\end{equation}

\subsection{Numerical Association Rule Mining}

NARM came into the scenario to extract association rules from numerical data.
Unlike the classical ARM, numerical ARM allows attributes to be categorical (e.g., gender, education) or numeric (e.g., salary, age) rather than just Boolean. A numerical association rule is an implication of the form $X \Rightarrow Y$, in which both antecedent and consequent parts are the set of attributes in the forms 
$A = \{v_1, v_2, \ldots v_n\}$ if A is a categorical attribute, or $A \epsilon~ [v_1,v_2]$ if A is numeric attribute.

An example of a numerical association rule is given below. $$ Age \in [21,35] \wedge Gender:[Male] \Rightarrow \textit{Salary} \in [2000,3000]$$    
$$\textit{(Support}=10\%, \textit{Confidence}=80\%)$$


This rule states that those employees who are males, aged between $21$ and $35$ and having salaries between \$2,000 and \$3,000 form $10\%$ of all employees; 
and that $80\%$ of males aged between $21$ and $35$ are earning between \$2,000 and \$3,000. Here, \emph{Age} and \emph{Salary} are numerical attributes and \emph{Gender} is a categorical attribute. In ARM, except for support and confidence, more than fifty measures of interestingness are available in the literature \cite{Hamilton2006,Rahulmeasures}. The support of an association rule $X \Rightarrow Y $  determines how frequently the itemset appears in a transactional database. The confidence of an association rule determines how many transactions that contain $X$ also contain $Y$.


\subsection{Related Work}
In recent years, there have been few surveys and studies in the literature that have focused on NARM approaches and their comparison. However, no SLR has been published to date. Our automated search identified three reviews \cite{adhikary2015trends,kaushik2020potential,kaushik2021systematic} and a manual search found two surveys \cite{adhikary2015mining, gosain2013comprehensive}. While these reviews provide a contribution towards understanding the methods and algorithms for NARM, they have several limitations, as outlined in Table \ref{reviews}. Gosain et al. \cite{gosain2013comprehensive} presented a survey of association rules on quantitative data in 2013. The authors focused on different types of association rules but did not include NARM methods and algorithms. Adhikary and Roy\cite{adhikary2015mining} reviewed QARM techniques, with a focus on applications in the real world, while their 2015 survey \cite{adhikary2015trends} provided a classification of QARM techniques but lacked valuable information. 
A systematic assessment of the three popular methods for NARM with thirty algorithms was conducted in \cite{kaushik2021systematic}. This review focused only on NARM algorithms, and the steps of systematic reviews were not followed. 
In contrast, our study conducted an SLR under the guidelines of Kitchenham and Charters \cite{article} and answered the research questions in the state of the art of NARM. 

Moreover, it is worth noting that some authors have made notable contributions to NARM under alternative names. For example, Telkani et al. \cite{telikani2020survey} conducted an extensive survey on evolutionary computation for ARM, wherein they thoroughly examined various approaches within the realm of ARM, including NARM, and provided insights into the 
the classification of evolutionary algorithms in this context.

\section{Research Methodology}
\label{methodology}
In this work, we adopt research methodology based on Kitchenham and Charters's guidelines \cite{article}. The main goal of this SLR is to summarize the existing evidence in the literature regarding this topic and to identify gaps in the literature. According to Kitchenham's guidelines, the process included three main phases: planning, conducting, and reporting the review. The planning phase involved identifying the need for the review and establishing a review protocol. The conducting phase involved following the review protocol, which included selecting a primary search, assessing the quality of the studies, and extracting relevant data. Finally, the reporting phase focused on formatting and evaluating the report in accordance with the guidelines.

\subsection{Planning the Review}
The initial phase of this study aims to justify the need for an SLR and define the research questions. Based on the objective and motivation of this study, we formulated the following research questions with the goal presented in Table \ref{RQs}. The primary aim of this SLR is to address these research questions, which will help us to comprehensively understand the existing research and identify gaps in the literature related to NARM. 

\begin{table}[!ht]
    \centering
    \caption{Research Questions Together With Their Goals}
    \label{RQs}
    \begin{tabular} {p{0.8cm} p{5cm} p{7cm}}
    \hline
ID & Research Question & Goal\\
\hline 
RQ1 & Which methods exist for solving NARM problems?   & Identifying NARM methods used by researchers in the literature to solve the NARM problem.\\
RQ2 & What are the several algorithms available for each of the existing NARM methods? &  Investigating state-of-the-art algorithms proposed under different NARM methods. \\ 
RQ3 &  What are the advantages and limitations of the existing NARM methods? & Exploring the benefits and limitations of existing NARM methods, along with their classification.\\
RQ4 & Which objectives are considered by the several existing multi-objective optimization NARM algorithms? & Providing an understanding of the objectives used for the multi-objective NARM algorithms under the optimization method.\\ 
RQ5 & What are the metrics to evaluate the NARM algorithms? & Discussing the metrics that have been used to evaluate algorithms and which ones are the most popular.\\
RQ6 & Which datasets are used for experiments by NARM methods? & Providing a detailed understanding of the datasets used for NARM methods.  \\
RQ7 & What are potential future perspectives for the area of NARM? & Discussing the  research challenges and future prospects that will help the researchers in future investigations and perform meaningful research.\\
RQ8 & How to automate discretization of numerical attributes for NARM in a useful (natural) manner? & Presenting an automated measure to discretize numerical attributes, which is particularly natural, i.e.,  which particularly well meets human perception of partitions.\\
\hline  
     \end{tabular}
\end{table}







\subsection{Conducting the Review}
The review phase involves a series of sequential steps, beginning with the identification of relevant research and followed by the selection of studies, study quality assessment, and data extraction.
These steps are conducted systematically to ensure the comprehensive coverage of relevant studies and the extraction of accurate and reliable information for analysis.

\subsubsection{Search Strategy}

\paragraph{Academic Databases}


To conduct the review phase, we conducted a thorough search of scientific publications from relevant journals and conferences, utilizing multiple reputable digital libraries, including the ACM Digital Library, Scopus, SpringerLink, IEEE Xplore, and ScienceDirect. Additionally, we performed a manual search on Google Scholar to minimize the chance of overlooking any significant articles. The search was conducted between April and June 2022, focusing on articles published in journals and conferences. We set the time frame for articles published from 1996 to 2022, as it was in 1996 when Srikant and Agrawal \cite{srikant1996} first presented the problem statement concerning numerical attributes.


\paragraph{Search Strings} For the search process, we derived the search terms from the research questions and compiled a comprehensive list of synonyms, abbreviations, and alternative words. In this study, we have also mentioned that the problem of handling numerical attributes was initially addressed as ``quantitative association rule'' by Srikant and Agrawal \cite{srikant1996}. Over time, this term evolved into ``numerical association rules.'' Therefore, to ensure inclusivity, our search terms included variations such as "quantitative association rule mining," OR "numerical association rule mining," OR "quantitative association rules," OR 
 "numerical association rules," OR  "quantitative ARM," OR "numerical ARM," OR "QARM," OR "NARM." 
We targeted these terms in the abstracts, titles, and keywords of articles within the following electronic sources.
\begin{itemize}
    \item ACM Digital Library\footnote{\url{http://dl.acm.org}}
    \item  IEEE eXplore\footnote{\url{http://ieeexplore.ieee.org}}
\item  Scopus\footnote{\url{http://www.scopus.com}}
\item  SpringerLink\footnote{\url{http://www.link.springer.com/}}
\item ScienceDirect\footnote{\url{https://www.sciencedirect.com/}}
\item Google Scholar\footnote{\url{https://scholar.google.com/}}
\end{itemize}


\begin{table}[!ht]
    \centering
    \caption{Search Terms}
    \label{Searchterm}
    \begin{tabular} {p{1.5cm}  p{12cm}}
    \hline
     Search Term &   "Quantitative Association Rule Mining" OR "Numerical Association Rule Mining" OR "Quantitative Association Rules" OR "Numerical Association Rules" OR "Quantitative ARM" OR "Numerical ARM" OR "QARM" OR "NARM" \\
     \hline
    Search String   & 
(TITLE-ABS-KEY ("Quantitative Association Rule Mining")  
  OR TITLE-ABS-KEY ("Numerical Association Rule mining")  
  OR TITLE-ABS-KEY ("Numerical Association Rule") 
  OR TITLE-ABS-KEY ("Quantitative Association Rule") 
  OR TITLE-ABS-KEY ("Quantitative ARM")  
  OR  TITLE-ABS-KEY ("Numerical ARM")  
  OR  TITLE-ABS-KEY ("QARM")  
  OR  TITLE-ABS-KEY ("NARM"))  
AND \mbox{PUBYEAR $>$ 1995}  
AND PUBYEAR  $<$ 2023  
AND LIMIT-TO (PUBSTAGE,"final")  
AND LIMIT-TO (SUBJAREA,"COMP")  
AND LIMIT-TO (LANGUAGE, "English")   \\
\hline

    \end{tabular}

\end{table}

\paragraph{Search Process}
Our search was specifically conducted for articles written in English, limited to the period between 1996 and 2022, within the subject area of Computer Science, focusing on the final publication stage. The search query and terms used in Scopus are outlined in Table \ref{Searchterm}.
Through a meticulous search process, we successfully identified a total of 1,628 articles. Following the elimination of 488 redundant articles, we narrowed down the selection to 1,140 articles. Table \ref{Searchresult} provides a breakdown of the number of articles obtained from each respective database.

\begin{table}[!ht]
    \centering
    \caption{Search Results from the  Digital Libraries}
    \label{Searchresult}
    \begin{tabular} {p{4cm}  p{3cm}}
    \hline
  Digital Library & Number of Results \\\hline
IEEE Xplore & 102\\
Scopus & 223 \\
SpringerLink & 618 \\
ACM & 187\\
ScienceDirect & 148\\
Google Scholar & 350 \\
\hline
Total & 1,628\\
\hline 
Redundant Articles & 488\\\hline 
Non-redundant Articles & 1,140\\
\hline  
  \\

    \end{tabular}

\end{table}


\subsubsection{Selection Based on Inclusion and Exclusion Criteria}
To ensure the relevance of the articles, we conducted an initial screening process by carefully reviewing the abstracts and conclusions. We applied the predetermined \emph{Inclusion and Exclusion Criteria}, which are outlined in Table \ref{IEcriteria}. These criteria are widely accepted and primarily focus on aligning with the scope of the study.

Non-peer-reviewed articles, such as theses and abstracts, were excluded from our analysis. Additionally, we also excluded works that combined results from both journals and conferences, such as monographs and books. Following the application of these inclusion and exclusion criteria, we were left with a final set of 96 articles that met our selection criteria. Next, to ensure a comprehensive review, we conducted a thorough examination of the references cited in the selected primary studies. This step aimed to identify any significant publications that might have been missed during the initial search. As a result, we identified 14 additional papers that fulfilled our inclusion criteria. These studies were subsequently incorporated into our list of primary studies, expanding the total number of articles to 110.


\begin{table}[!ht]
    \centering
    \caption{Inclusion and Exclusion  Criteria }
    \label{IEcriteria}
    \begin{tabular} {p{1.5cm}  p{7.5cm}}
    \hline
ID &Inclusion Criteria \\\hline
I1 & The article discusses a novel method for mining the quantitative or numerical association rules.\\
I2 & The article proposes a novel algorithm for mining the quantitative or numerical association rules.\\
I3 & The article discusses an extension to the existing algorithm for mining the quantitative or numerical association rules.\\
I4 & The article is related to at least one of the proposed research questions. \\
I5 & The article describes the theoretical foundation of mining the association rules from numerical data sets.\\

\hline
 ID & Exclusion Criteria \\ \hline 
E1 & Articles which are only application-oriented.\\
E2 & Articles present surveys and short papers.\\
E3 & Abstracts, editorials, thesis, monographs, panels, books.\\
E4 & Conference version of an article whose journal version is included.\\
\hline
\end{tabular}

\end{table}

\subsubsection{Selection based on Quality Assessment}
The objective of the quality assessment phase is to ensure the inclusion of unbiased and relevant studies in the review. To accomplish this, we established a set of criteria to evaluate the quality of the papers, refine our search results, and assess the relevance and rigour of the included papers. Following the initial selection based on the predefined inclusion and exclusion criteria, we conducted a thorough reading of the entire article. During this phase, we utilized a quality assessment checklist comprising five criteria, as outlined in Table \ref{QAchecklist}, to refine our search results. Each criterion was evaluated using ``Yes,'' ``No,'' or ``Partially'' responses, which corresponded to scores of 1, 0, or 0.5, respectively. Articles with scores of 2.5 or higher were selected as the final primary studies. Through this rigorous quality assessment process, we determined a total of 68 articles that met our selection criteria and were deemed as the final primary studies. The list of final articles is available in the GitHub repository\footnote{\url{https://github.com/minakshikaushik/List-of-Final-selected-articles.git}}. 


\begin{table}[!ht]
    \centering
    \caption{Quality Assessment Checklist}
    \label{QAchecklist}
    \begin{tabular} {p{1.5cm}  p{7.5cm}}
    \hline
 ID & Quality Questions \\\hline
QQ1 & Are the proposed methods in the articles well defined?\\
QQ2 & Are the methods/algorithms/experiments defined clearly? \\
QQ3 & Are the results validated? \\
QQ4 & Are their any solid finding/result and clear outcomes?\\
QQ5 & Is the contribution of the article clearly defined?\\

\hline

\hline
\end{tabular}
\end{table}

\subsubsection{Data Extraction and synthesis}

In the last phase, we extracted pertinent information from the selected articles that successfully passed the quality assessment. This information was utilized to generate a comprehensive summary of our findings. Each chosen article was downloaded and thoroughly examined. Table \ref{Dataextraction} provides an overview of the extracted data from each publication, highlighting its relevance to the respective research questions. For a more in-depth analysis of the collected data and the synthesis of our findings, we encourage readers to refer to Sections \ref{result} and \ref{discussion}. These sections provide a detailed presentation of the information gathered from the final set of articles, offering valuable insights into the research questions and facilitating a comprehensive understanding of our review's outcomes.

\begin{table}[!ht]
    \centering
    \caption{Data Extracted from Selected Articles Based on Our Research Questions}
    \label{Dataextraction}
    \begin{tabular} {p{7cm}  p{2cm}}
    \hline
 Extracted Data & Related RQ \\\hline
Article's title & General\\
Author's name & General \\
Source name & General\\
Type of publication (conference/Journal) & General\\
Year of publication & General\\
Citation count & General\\
Methods & RQ1, RQ3 \\
Algorithms & RQ2\\
Name of datasets & RQ6\\
Source of datasets & RQ6\\
Objectives & RQ4\\
Metrics & RQ5\\
\hline
\end{tabular}
\end{table}

\section{Reporting the Review}
\label{result}

The reporting phase is crucial as it involves the final presentation and evaluation of the findings obtained from the systematic review. Effectively communicating the results is essential to highlight the contribution of the review and provide valuable insights to readers. These results are derived from the studies identified during the review phase and are aligned with the pre-defined research questions. Through clear and concise reporting, the systematic review aims to enhance understanding and facilitate informed decision-making.

\subsection{RQ1.Which methods exist for solving NARM problems?}
\label{RQ1}

The selected studies, which are reviewed to examine the existing methods in NARM, are summarized in the subsequent sub-sections. Table \ref{Overview} provides an overview of the included papers pertaining to different NARM methods. Following a thorough analysis of these studies, it was determined that they could be broadly categorized into four main methods. The following subsections provide brief descriptions of these methods.


\subsubsection{The Discretization Method}
\label{TheDiscretizationMethod}
Classical ARM faces a significant limitation when dealing with continuous variable columns as they cannot be processed directly and must be converted into binary form first. To address this issue, researchers have turned to the discretization method \cite{mg2000,khade2015supervised,moreland2009discretization}. Discretization involves dividing a column of numeric values into meaningful target groups, which facilitates the identification and generation of association rules. This approach helps to understand numeric value columns easily, but the groups are only useful if the variables in the same group do not have any objective differences. Additionally, discretization minimizes the impact of trivial variations between values. The discretization method for mining numerical association rules can be categorized into four approaches: partitioning, clustering,  fuzzifying and hybrid. In this article, we have selected 28 relevant studies that focus on the discretization method.

\paragraph{Partitioning Approach}
Srikant \cite{srikant1996} presented a solution for mining association rules from quantitative data sets. The approach involved partitioning the numerical attributes into intervals and subsequently mapping these intervals into binary attributes. To address the information loss resulting from partitioning, the authors introduced the concept of the \emph{partial completeness measure}. By partitioning the numerical attributes and mapping them into binary attributes, Srikant's approach allowed for the application of traditional ARM techniques to quantitative data. This work laid the foundation for handling numerical attributes in ARM and has since influenced further developments in the field.

\paragraph{Clustering Approach} 
The clustering approach is utilized to divide a numerical column into distinct groups based on similarity among values. Various clustering techniques, including merging-based, density-based, and grid-based clustering, can be employed to achieve this goal. From the clustering approach, we identified nine relevant articles that explore this methodology.

In the merging and splitting-based concept, intervals are merged initially and then subsequently split based on specific criteria. Wang and Han proposed the notion of merging adjacent intervals in their work \cite{wang1998interestingness}. Li et al. \cite{li} developed a method that identifies intervals of numeric attributes and merges adjacent intervals exhibiting similar characteristics based on predefined criteria. These studies contribute to the understanding and advancement of the merging and splitting-based approach within the context of NARM.

The density-based clustering aims to identify different dense regions within the dataset and map these regions to numeric association rules. Algorithms such as DRMiner \cite{lian2005efficient}, DBSMiner \cite{Guo2008}, and MQAR \cite{Yang2010} are examples of techniques proposed within this category. Further details regarding these algorithms will be provided in response to the subsequent research question. On the other hand, grid-based clustering utilizes a bitmap grid to handle data clustering. It identifies clusters within the bitmap grid, which subsequently yield association rules. This method offers an alternative approach for extracting meaningful associations from numerical attributes.

\paragraph{Fuzzy Approach}


The fuzzy approach is employed to tackle the issue of sharp boundaries in ARM by representing numerical values as fuzzy sets. Fuzzy sets allow for the representation of intervals with non-sharp boundaries, where an element can possess a membership value indicating its degree of belonging to a set. Hong et al. \cite{hong1999mining} applied the fuzzy concept in conjunction with the apriori algorithm to discover fuzzy association rules from a quantitative dataset. Their work demonstrated the effectiveness of combining fuzzy sets and ARM techniques for extracting valuable insights from numerical data.

\paragraph{Hybrid Approach}
The hybrid approach for solving NARM problems is the combination of two or more methods such as clustering, partitioning, and fuzzy approaches. This method is a more flexible approach that can enhance the efficiency and accuracy of ARM. For instance, \cite{zhang1999mining} combined the fuzzy approach with the partitioning method to develop an efficient algorithm for mining fuzzy association rules. On the other hand, \cite{Mohamadlou2009, Tsmodel, kianmehr2010fuzzy} utilized the fuzzy approach with clustering to enhance the accuracy of ARM. The hybrid approach in NARM offers a promising direction for researchers to explore, as it allows for the utilization of complementary techniques to address the complexities of mining association rules from numerical data.

\begin{table}[!ht]
    \centering
    \caption{Overview of Solutions Based on NARM Methods}
    \label{Overview}
    \begin{tabular} {p{2.4cm} p{3.1cm} p{1.5cm}  p{5cm}}
\toprule
 Methods & Approaches & \# included papers & References \\
 \midrule
 \multirow{5}{*}{Discretization} & Partitioning & 9 & \cite{srikant1996,chan1997effective,buchter1998,fukuda1999mining,brin1999mining, rastogi2002mining, li,SWP2019,Song2013}
\\ 
  
 & Clustering & 9 & \cite{Miller1997,lent1997clustering,wang1998interestingness,mg2000,lian2005efficient,Guo2008,Yang2010,Dong2014,Medjadba2019}   \\
 
 & Fuzzy & 6  & \cite{chan1997mining,Kuok1998,hong1999mining,Gyenesei2001AFA,Lee2001,zheng2014optimized}  
\\
 & Hybrid  & 4   & \cite{zhang1999mining,Mohamadlou2009,Tsmodel,kianmehr2010fuzzy}
\\
\midrule
\multirow{9}{*}{Optimization} & Evolutionary &  17 & \cite{mata2001mining,mata2002discovering,alvarez2012evolutionary,alatacs2006efficient,salleb2007quantminer,yan2009genetic,martinez2010mining,martinez2011evolutionary,MARTIN2016,MARTIN2014,MARTIN2014mopnar,martinez2016improving,MOEA-QAR,ALMASI2015,taboada2008association,luna2014reducing,Minaeibidgoli2013}\\

 & Differential Evolution & 3 & \cite{Fister2018,ALATAS2008modenar,ALTAY2021DEsine} \\

 & Swarm Intelligence & 11 & \cite{alatas2008rough,ALATASCENPSO,yan2019ppqar,beiranvand2014multi,tahyudin2019improved, wolf,kuo2019multi,moslehi2011multi,kahvazadeh2015mocanar,heraguemi2018mBAT,ledmi2021discrete}
\\
 & Physics-based & 1  & \cite{can2017automatic} \\

 & Hybrid & 2 & \cite{moslehi2020novel,altay2022chaos} \\

\midrule
Statistical &  & 3 & \cite{aumann2003statistical,kang2009bipartition,webb2001discovering} 
\\
\midrule
Other &  & 3 & \cite{ke2008information,hu2022cognitive,QMVMO}
\\
\bottomrule
\end{tabular}
\end{table}

\subsubsection{The Optimization Methods}
In the context of NARM, the optimization method has gained significant attention, and we identified 34 papers out of the 68 studies reviewed that focused on optimization methods. These methods utilize heuristic algorithms inspired by various natural phenomena, such as animal movements and biological behavior. Generally, optimization methods fall into two categories: bio-inspired and physics-based. Depending on the optimization goals, the optimization methods can be further classified into single-objective and multi-objective approaches. 

Bio-inspired optimization methods consist of approaches based on Swarm Intelligence (SI), Evolutionary algorithms, and Hybrid methods. These methods draw inspiration from the collective behavior of organisms in nature. For example, some studies have explored algorithms inspired by the movements of wolves \cite{wolf}, insects \cite{moslehi2011multi}, and mining behavior in biological systems \cite{mata2001mining}. The physics-based optimization methods apply principles from physics to solve optimization problems. These approaches offer researchers a diverse range of techniques to explore and apply in NARM, allowing for the discovery of efficient and effective association rules from numerical data.

\paragraph{Evolution-Based Methods}

The evolutionary method in NARM is rooted in Darwin's theory of natural selection, which highlights the adaptive nature of living organisms in response to changing environments. This approach employs biological operators, including crossover, mutation, and selection, to mimic the evolutionary process in optimization algorithms \cite{eiben2015}. By applying these principles, evolutionary methods aim to enhance the effectiveness and efficiency of NARM algorithms, allowing for the discovery of valuable association rules from numerical data.

Under the evolution-based method, the genetic algorithm (GA) and differential evolution (DE) provide detailed solutions for the NARM problem. The optimization method aims to discover association rules without the need for the prior discretization of numerical attributes.
GA, a meta-heuristic inspired by natural selection and genetic structure, evolves a population of individual solutions over time \cite{holland1992adaptation}. It proceeds in three main steps: selection of parent individuals, crossover to combine parents for the next generation, and mutation to apply random changes to parents and form children. In 2001, the concept of genetic algorithms was successfully applied to identify numerical association rules from numerical attributes \cite{mata2001mining}.

Out of the selected studies, $17$ refers to the use of genetic algorithms. Initially, NARM algorithms focused solely on single-objective problems; later, multi-objective algorithms also came into the scenario \cite{Minaeibidgoli2013}. Over the years, the genetic algorithm has been used with some advancement by integrating various supporting techniques, such as the binary-coded CHC algorithm \cite{martinez2011evolutionary}, non-dominated sorting genetic algorithm \cite{MARTIN2014}, and niching genetic algorithm \cite{MARTIN2016}, as well as other multi-objective genetic algorithms. Genetic programming \cite{koza1994genetic}, which utilizes a tree structure for the genome, is another aspect of the genetic algorithm. Grammar-guided genetic programming \cite{luna2014reducing,luna2016mining} also emerged with NARM in 2004.

In 1997, Storn and Price \cite{storn1997differential} introduced a global optimization meta-heuristic approach that effectively minimized non-differentiable, non-linear, and multi-modal cost functions. This approach utilized the same operator as genetic algorithms, which included crossover, mutation, and selection. To minimize the function, differential evolution (DE) employed a few control variables and parallelization techniques, which helped to decrease computing costs and quickly converge on the global minimum. Our research identified four relevant studies that used DE for NARM. One such study, proposed in 2008 by Alatas and Akin, utilized a multi-objective differential evolution algorithm \cite{ALATAS2008modenar}. Another study was conducted in 2018 and 2021 by I. Fister Jr. \cite{Fister2018, FisterImproved}, while Altay and Alatas presented a hybrid DE-based method with a sine cosine algorithm and chaos number-based encoding, respectively \cite{ALTAY2021DEsine, altay2022chaos}.

\paragraph{Swarm Intelligence-Based}
Swarm intelligence (SI) is a popular optimization technique inspired by the collective behavior of self-organized groups in nature, as described by Bonabeau et al. in 1999 \cite{bonabeau1999swarm}. SI algorithms emulate the behavior of swarms found in birds, fish, honey bees, and ant colonies. These algorithms consist of individuals that migrate through the search space, simulating the progression of the swarm. Various SI-based algorithms have been developed, including Particle Swarm Optimization (PSO), Bat Algorithm (BAT), Ant Colony Optimization (ACO), Cat Swarm Optimization (CSO), and others. In the context of solving NARM problems, several SI algorithms have been applied. Notable examples include PSO \cite{alatas2008rough}, BAT \cite{heraguemi2018mBAT}, Wolf Search Algorithm (WSA) \cite{wolf}, Crow Search Algorithm (CSA) \cite{ledmi2021discrete}, and Cuckoo Search Algorithm (CS)\cite{kahvazadeh2015mocanar}. These SI-based algorithms have shown promise in optimizing NARM and extracting meaningful association rules.


Particle swarm optimization (PSO) is a widely used optimization technique for non-linear continuous functions inspired by the movement of bird flocks or fish schools as described in Kennedy and Eberhart \cite{kennedy1995particle}. PSO simulates the collective behaviour of these groups, where $N$ particles move in a $D$-dimensional search space, adjusting their position iteratively by using their own best position \emph{pbest} and the best position of the entire swarm \emph{gbest}. The PSO algorithm finds the optimum solution by calculating the velocity and position of each particle. In the context of mining association rules with numeric attributes, Alatas and Akin introduced the application of PSO in 2008 \cite{alatas2008rough}. They modified the PSO algorithm to search for numeric attribute intervals and discover numeric association rules. Seven studies have since focused on adapting PSO for NARM including the hybrid approach. These studies explore the potential of PSO to effectively mine association rules with numeric attributes and provide valuable insights into its performance and limitations.

Ant colony optimization (ACO) is another optimization technique based on the foraging behaviour of various ant species, as described in Dorigo et al.\cite{dorigo2006ant}. In ACO, a group of artificial ants collaborates to find solutions to an optimization problem and communicate information about the quality of these solutions using a communication mechanism similar to real ants. ACO is designed to address discrete optimization problems by selecting a solution using a discrete probability distribution. In the context of multi-objective NARM, Moslehi et al. introduced an ACO variant called $ACO_R$ in 2011 \cite{moslehi2011multi}. $ACO_R$ utilizes a Gaussian probability distribution function to handle continuous values encountered in NARM. It maintains a solution archive of size $k$, initially populated with $k$ random solutions ranked by their quality. Each ant constructs its solution by probabilistically selecting a solution from the archive, allowing for the exploration of different solution possibilities. The utilization of ACO in NARM, particularly the $ACO_R$ variant, demonstrates its potential to address the challenges posed by continuous attributes and provide effective solutions for multi-objective NARM problems.

The Cuckoo Search algorithm (CS) is an optimization algorithm introduced by Yang and Deb in 2009, inspired by the brooding parasitic behavior of cuckoo species \cite{yang2009cuckoo}. Cuckoos lay their eggs in the nests of other bird species, mimicking the color and pattern of the host birds' eggs. Some host birds may recognize the stranger's eggs and remove them from the nest. The cuckoo search algorithm mimics this behavior by generating new solutions (cuckoo eggs) and replacing less promising solutions in the nests (solution space) with the new solutions. The algorithm operates based on three main rules: A cuckoo bird lays only one egg at a time in a randomly chosen nest (introduces a new solution to the search space). The nests with high-quality eggs are more likely to be carried over to the next generation (the better solutions have a higher chance of survival). The probability of a host bird discovering cuckoo eggs in its nest is either 0 or 1 (either the host bird finds and removes the cuckoo egg or it remains undetected). The goal of the cuckoo search algorithm is to find new and potentially better solutions to replace the existing solutions in the nests, leading to the improvement of the overall solution quality. In the context of NARM, a multi-objective cuckoo search algorithm called MOCANAR was proposed by Kahvazadeh et al. in 2015 \cite{kahvazadeh2015mocanar}. MOCANAR applies a Pareto-based approach to solve the multi-objective NARM problem, aiming to discover association rules that optimize multiple conflicting objectives simultaneously. By employing the cuckoo search algorithm as the underlying optimization technique, MOCANAR demonstrates its effectiveness in addressing the challenges of multi-objective NARM.

In 2012, Tang et al. proposed a heuristic optimization algorithm called the Wolf Search Algorithm (WSA) that imitates how wolves hunt for food and survive in the wild by avoiding predators \cite{tang2012wolf}. Unlike other bio-inspired meta-heuristics, WSA enables both individual local searching and autonomous flocking movement capabilities as wolves hunt independently in groups. WSA follows three basic rules based on wolf hunting behavior. The first rule involves a fixed visual area of each wolf with a radius \emph{v}, which is calculated using Minkowski distance. The second rule pertains to the current position of the wolf, represented by the objective function's fitness, and the wolf always tries to choose the better position. The third rule concerns escaping from enemies. Agbehadji suggested WSA to develop an algorithm for searching for intervals of numeric attributes and association rules \cite{wolf}.

In 2010, Yang introduced the BAT algorithm (BA) as a solution to continuous constrained optimization problems inspired by the echolocation behavior of microbats \cite{yang2010BAT}. Microbats use echolocation to sense distance, discover prey, avoid obstacles, and find roosting nooks in the dark. The BA algorithm is based on the velocity of a bat at a particular position, with a fixed frequency and varying wavelength and loudness. The bat adjusts its frequency and loudness to locate a new food source while changing its position in space. Heraguemi et al. \cite{heraguemi2018mBAT} proposed a multi-objective version of the Bat algorithm for numerical attributes. Previously, the BA was also used for ARM to deal with categorical attributes.

The Crow Search Algorithm (CSA) is a recently developed meta-heuristic optimization technique inspired by the intelligent behaviour of crows \cite{ASKARZADEH20161}. Crows are known for their ability to store and hide food for future use while also keeping an eye on each other to steal food. The CSA is based on four principles of crow behaviour: living in flocks, memorizing the position of hiding places, following other crows to steal food, and protecting their caches from theft. In the CSA, a crow flock moves in a $d$-dimensional search space, with each crow having its own position and memory of its hiding place. When a crow follows another crow, it may either discover the hiding place and memorize it or be tricked by the followed crow. The CSA has been successfully applied to various optimization problems, such as image segmentation and feature selection. Recently, Makhlouf et al. (2021) \cite{ledmi2021discrete} proposed a discrete version of CSA for NARM.

\paragraph{Hybrid Approach}
The hybrid approach in optimization combines multiple techniques such as evolution, SI, or other approaches to leverage their respective advantages and tackle complex tasks effectively. In the context of NARM, researchers have explored the hybridization of different algorithms to enhance the performance and efficiency of association rule discovery.
One study by Moslehi et al. \cite{moslehi2020novel} employed a hybrid approach that combined the GA and PSO. The GA facilitated the search for the best solution, while the PSO helped avoid being trapped in local optima by exploring a larger search space. By combining the strengths of both approaches, the hybrid algorithm demonstrated the ability to find high-quality solutions to complex NARM problems within a relatively short time. Another study by Altay and Alatas \cite{altay2022chaos} proposed a hybrid approach that combined the DE algorithm with the sine and cosine algorithms. This hybridization aimed to leverage the exploration and exploitation capabilities of both algorithms, resulting in improved performance for NARM. The DE algorithm provided efficient search and optimization, while the sine and cosine algorithms introduced chaos-based techniques to enhance the exploration process.


\paragraph{Physics-Based} 
Physics-based meta-heuristics have emerged as a powerful approach to solving optimization problems. One such algorithm, the gravitational search algorithm (GSA) \cite{rashedi2009gsa}, is based on Newton's law of gravity, where particles attract each other with a gravitational force.
The following formula defines this force: 
\begin{equation}
\label{eq3}
    F= G {\frac{M_1 M_2} {R^2}}
\end{equation}
where $F$ is the gravitational force, $G$ is the gravitational constant, $M_1$ and $M_2$ are the mass of of two particles and $R$ is the distance between these particles. According to Newton's second law, when a force is applied to a particle, its acceleration $a$ depends on the force $F$, and it is mass $M$.
\begin{equation}
\label{eq4}
    a= \frac{F} {M}
\end{equation}

In the GSA, agents are considered as objects with masses that determine their performance. The heavier masses are better solutions and attract lighter masses, leading to an optimal solution. Each mass has a position, inertial, active, and passive gravitational mass. The position of a mass represents a problem solution, and its gravitational and inertial masses are calculated using a fitness function.
While GSA has been used in various optimization problems, it has only been applied to NARM in one study, where Can and Alatas utilized it for finding intervals of numeric attributes automatically without any prior processing \cite{can2017automatic}.

\subsubsection{The Statistical Method}
Statistics is a traditional approach for developing theories and testing hypotheses using statistical tests such as Pearson correlation, regression, ANOVA, t-test, and chi-square test, among others. Statistical inference involves inferring population properties from a sample to generate estimates and test hypotheses.
Some studies have used statistical concepts such as mean, median, and standard deviation in the mining association rule. We identified three studies in this direction which suggested distribution-based interestingness measures. 
One such study is Kang et al. (2009) \cite{kang2009bipartition}, which used bipartition techniques such as mean-based bipartition, median-based bipartition and standard deviation minimization for quantitative attributes in ARM.

\subsubsection{Miscellaneous Other Methods}

In addition to the established techniques discussed earlier, there are other alternative approaches that have been proposed to tackle the challenge of NARM. These approaches offer unique perspectives and methodologies to address the problem. One such approach is the utilization of mutual information, as presented by Yiping et al. in 2008 \cite{ke2008information}. Mutual information is a concept from information theory that measures the dependency between two variables. In the context of NARM, mutual information is employed to generate quantitative association rules (QARs), capturing the relationships and dependencies between numerical attributes. Another approach is the use of Variable Mesh Optimization (VMO), proposed by Jaramillo et al. \cite{QMVMO}. VMO is a population-based metaheuristic algorithm that represents solutions as nodes distributed in a mesh-like structure. Each node in the mesh represents a potential solution to the optimization problem. By leveraging the principles of VMO, the algorithm explores the solution space in a distributed and adaptive manner, facilitating the discovery of association rules. Furthermore, in 2021, Hu et al. \cite{hu2022cognitive} introduced a cognitive computing-based approach for NARM. Cognitive computing refers to the simulation of human thought processes by computer models. By leveraging cognitive computing techniques, the proposed approach aims to mimic the human thought process during critical situations, allowing for a more comprehensive and nuanced analysis of numerical data for ARM.

These alternative approaches demonstrate the diverse range of methodologies and concepts that researchers have explored to tackle the NARM problem. By leveraging mutual information, variable mesh optimization, and cognitive computing, these approaches offer unique perspectives and potential benefits for discovering association rules from numerical data.

\subsection{RQ2 What are the several algorithms available for each of the existing NARM methods? }
In response to RQ1, we have provided a comprehensive explanation of the four main methods utilized in NARM in subsection \ref{RQ1}. This section further delves into a more detailed exploration of the algorithms associated with each of these methods.

\subsubsection{The Discretization Method}
\label{TheDiscretizationMethodALGOS}

\paragraph{Partitioning Based Algorithms}
\begin{itemize}
\item \emph{Qunatitative Association Rule Mining (QARM):}
In 1996, Srikant and Agrawal proposed an algorithm \cite{srikant1996} to address the use of numeric attributes in ARM, which was traditionally limited to binary attributes. One key issue was determining whether and how to partition a quantitative attribute while minimizing information loss by setting minimum support and confidence thresholds. To overcome this, the algorithm introduces a \emph{partial completeness measure}. The algorithm converts categorical attributes to integers and partitions numerical attributes into intervals using an equi-depth discretization algorithm. Frequent itemsets are then generated by setting minimum support for each attribute and used to generate association rules. To ensure interesting and non-redundant rules, the algorithm employs an interesting measure called ``greater-than-expected-values.'' However, setting the user-supplied threshold too high can result in missed rules, while setting it too low can generate irrelevant rules.

\item \emph{Automatic Pattern Analysis and Classification System 2 (APACS2):} 
To address the threshold issue, a novel algorithm named APACS2 was presented by Chan et al. \cite{chan1997effective}. This algorithm employed equal-width discretization to discover intervals of quantitative attributes without the need for user-defined thresholds. The quantitative attribute values were mapped to these intervals to obtain a new set of attributes. Each interval was described by the lower and upper bounds as $a_1 = [l_1, u_1]$. The APACS2 algorithm used \emph{adjusted difference} analysis to identify interesting associations between items, which enabled it to generate both positive and negative association rules.

\item \emph{Q2:}
Buchter and Wirth \cite{buchter1998} proposed the \emph{Q2} algorithm to work with multi-dimensional association rules over ordinal data. \emph{Q2} aimed to reduce the cost of counting a large number of buckets by only counting the buckets of successful candidates. First, apriori is used to identify all frequent boolean itemsets. Then, only the items in these sets are discretized based on the user's specifications. Q2-gen technique is used to generate a prefix tree that includes only the bucket combinations that need to be counted for the discretized items. The prefix tree is then used to count these bucket combinations in a single pass through the data. Finally, the prefix tree is used to produce all R-interesting rules. Unlike the hash tree used in \emph{QARM}, \emph{Q2} uses a prefix tree to store quantitative itemsets.

\item \emph{Fukuda et al. Work:} 
Fukuda et al. presented a novel algorithm \cite{fukuda1999mining} that computes two optimized ranges for numeric attributes. To achieve this, the algorithm uses randomized bucketing as a preprocessing step to compute the ranges for sorted data. The focus of the algorithm is on generating optimized rules of the format $(A [v_1, v_1]) \wedge C_1 \Rightarrow C_2$, where $C_1$ and $C_2$ are binary attributes and $A$ is a numeric attribute. The main task of the algorithm is to generate thousands of equi-depth buckets and combine some of them to generate optimized ranges. The performance of the bucketing algorithm was compared with Naive Sort and Vertical Split Sort, and the algorithm demonstrated superior performance.

\item \emph{Brin's Algorithm:} 
In 1999, Brin et al. proposed an optimized algorithm for mining one and two numeric attributes \cite{brin1999mining}. The focus of the algorithm was on optimizing gain rules, where the gain of a rule $R$ is defined by the difference between the support of $(antecedent \wedge consequent)$ and the support of $antecedent$, multiplied by the user-specified minimum confidence. To reduce the input size, a bucketing algorithm was employed. For one numeric attribute, the algorithm computes optimized gain rules, while for two numeric attributes, a dynamic programming algorithm was presented to compute approximate association rules. Although the algorithm was successful for one numeric attribute, it was not well-suited for large domain sizes in the case of two numeric attributes.

\item \emph{Numerical Attribute Merging Algorithm:}
Li et al. \cite{li} developed an algorithm that merges adjacent intervals of numeric attributes based on a merging criterion that considers value densities and distances between values. They called this the \emph{numerical attribute merging algorithm} and used it to find suitable intervals for the QARM algorithm. After discretizing the numeric attributes, this algorithm treats each interval as a boolean attribute, allowing them to work with classical ARM.

\item \emph{Rastogi's Algorithm:}
In 2002, Rastogi and Shim extended the work done by \cite{rastogi2002mining}. They presented efficient methods for reducing the search space during the computation of optimized association rules applicable to both categorical and numeric attributes.

\item \emph{Sliding Window Partitioning - Random Forest (SWP-RF) Algorithm:}
In a related study, Guanghui Fan et al. \cite{SWP2019} proposed a machine learning-based QARM method called \emph{SWP-RF} to identify factors that cause network deterioration. This method uses sliding window partitioning (SWP) to discretize continuous attributes into boolean values, followed by random forest (RF) feature importance to measure the association between key performance indicator (KPI) and key quality indicator (KQI).
 
\item \emph{Numerical Association Rule-Discovery:}  
Song and Ge \cite{Song2013} proposed NAR-Discovery, a divide-and-conquer algorithm for mining numerical association rules. NAR-Discovery progresses in two phases. In the first phase, attributes are partitioned into a small number of large buckets, and then neighbouring buckets are mapped to an ``item,'' and apply a classical frequent itemset mining algorithm. In the second phase, only the outermost buckets of each rule are recursively partitioned, and some bounds and filtering are used to end the process. The authors improved performance by one to two orders of magnitude using optimization techniques. They developed a search based on a tree structure to manage rule derivations, and interesting rules were selected using an optimization technique based on temporary tables. NAR-Discovery was compared with QuantMiner \cite{salleb2007quantminer} and claimed to discover all appropriate rules.

\end{itemize}

\paragraph{Clustering Based Algorithms}

\begin{itemize}
  \item \emph{Miller's Algorithm:}
Miller et al. \cite{Miller1997} introduced a distance-based ARM approach for interval data in 1997. To handle the memory requirements, they utilized a $B^+$ tree data structure. The authors first used a clustering algorithm to identify intervals and then applied a standard ARM algorithm to extract association rules from these intervals.

 \item \emph{Association Rule Clustering System (ARCS):}
In 1997, Lent et al. \cite{lent1997clustering} introduced a comprehensive framework called ARCS that focused on rules with two quantitative attributes on the antecedent side and one categorical attribute on the consequent side. ARCS consists of four main components: binner, association rule engine, clustering, and verifier. In the binner phase, quantitative attributes are divided into bins using the equi-width binning method, and these bins are then mapped to integers. The BitOp algorithm is used to enumerate clusters from the grid and locate them within the Bitmap grid by performing bitwise operations, which results in clustered association rules. However, this method is limited to handling low-dimensional data and cannot handle high-dimensional data.

 \item \emph{Interval Merger Algorithm:}
In 1998, Wang and Han \cite{wang1998interestingness} proposed an algorithm for merging adjacent intervals of numeric attributes by evaluating merging criteria. This algorithm has two phases: initialization and bottom-up merging. They used an $M$-tree, which is a modified $B$-tree, to efficiently find the best merge during the merging phase. Additionally, two interestingness measures, $J_1$ and $J_2$, were used to evaluate the interestingness of the discovered association rules. The higher the values for both measures, the more interesting the rule was considered to be.

 \item \emph{Relative Unsupervised Discretization (RUDE):}
In 2000, Ludl et al. proposed the RUDE algorithm as a merging approach based on the merging and splitting technique \cite{mg2000}. The RUDE algorithm considers the interdependence of attributes and consists of three main steps. The first step is the pre-discretizing phase, where equal-width discretization is applied to the data. In the second step, called structure projection, the structure of each source attribute is projected onto the target attribute. This projection is then used to perform clustering on the target attribute, resulting in the gathering of split points in the split point list. Finally, in the postprocessing step, the split points are merged using predefined merging parameters. The RUDE algorithm was primarily used as a preprocessing step for the apriori algorithm. The association rules extracted from RUDE and apriori were combined to obtain the final results.


 \item \emph{Dense Regions Miner (DRMiner):}
In 2005, Lian et al. \cite{lian2005efficient} proposed the DRMiner algorithm, which efficiently identifies dense regions and maps them to QARs. To achieve this, the authors developed a three-step approach. First, a $k-d$ tree is built to store valid cells in the space and their corresponding number of points. Second, a dense region cover set is grown inside some leaf nodes from their boundaries, and self-merging of cover sets is done across boundaries. Finally, the cells are traversed in each cover to find dense regions. The authors evaluated the complexity of DRMiner for different steps and used a synthetic data set with varying numbers of attributes and instances for evaluation.

 \item \emph{Density-Based Sub-space
Miner (DBSMiner):}
The DBSMiner algorithm, proposed in 2008 by Guo et al. \cite{Guo2008}, aims to cluster the high-density subspace of quantitative attributes. CBSD (Clustering Based on Sorted Dense Units), a new clustering algorithm, was used to sort all subspaces with densities greater than a certain threshold in descending order. Interestingly, DBSMiner has a unique property when dealing with low-density subspaces: it only needs to verify the neighbouring cell instead of scanning the entire space. The algorithm is capable of uncovering interesting association rules.

\item \emph{Mining Quantitative Association Rule (MQAR):}
Yang et al. \cite{Yang2010} proposed the MQAR algorithm in 2010, which utilizes dense regions to generate numerical association rules. The algorithm clusters dense subspaces using the DGFP tree (dense grid frequent pattern tree) in four main steps. Firstly, the data space is partitioned into non-overlapping rectangular units by partitioning each quantitative attribute into intervals. Then, a DGFP tree is created to store dense cells in the space with a density greater than the minimal density criterion by mapping all database transactions into a high-dimensional space $S$ and sorting units by density. The third step is to mine the DGFP tree to obtain dense subspaces, which provide information about database transactions. Finally, the dense subspaces in $S$ are identified based on the dense subspaces, and associated cells are found to build clusters. Association rules that are not redundant are then constructed using the clustering result.

\item \emph{Quantitative Association Rule Mining Method with Clustering Partition (QARC\_Apriori):}
The QARC\_Apriori algorithm, proposed in 2014, aimed to analyze correlations in satellite telemetry data \cite{Dong2014}. The algorithm involved three main steps: First, it performed dimensionality reduction to eliminate redundant attributes. Second, it discretized numeric attributes using the $K$-means clustering algorithm. Finally, it used the apriori algorithm for mining QARs, with frequent itemset mining and rule generation. Since satellite telemetry data has a vast amount of data, numerical attributes, and high dimensions with various attributes such as voltage, current, pressure, and temperature, the authors used the grey relational analysis method to reduce the dimensionality.

\item \emph{Graph Clustering and Quantitative Association Rules (GCQAR):}
Medjadba et al. \cite{Medjadba2019} proposed GCQAR, a method for discovering significant patterns in geochemical data by combining graph clustering and QARM. Identifying hidden patterns related to mineralization in geochemical data is a challenging task. The proposed method tackles this by first applying graph clustering to partition the input data into highly cohesive, sparsely connected subgraphs. This step helps to separate the relevant geochemical data from the complex background. Then, QARs are used to measure the interrelation between pairs of vertices in each subgraph. For each cluster, a set of QARs is generated by randomly selecting antecedent and consequent rules and evaluating them based on support and confidence.
\end{itemize}

\paragraph{Fuzzy Based Algorithms}
\begin{itemize}
 \item \emph{Fuzzy-Automatic Pattern Analysis and Classification System (F-APACS):}
Chan extended the APACS2 algorithm for QARM by proposing the F-APACS algorithm \cite{chan1997mining}, which is based on fuzzy set theory and is designed for mining association rules with numeric attributes. Instead of finding intervals for quantitative attributes as done in other methods, F-APACS uses linguistic terms to represent discovered patterns and exceptions.
Similar to APACS2, F-APACS also employs the \emph{adjusted difference analysis} technique, which eliminates the need for a user-supplied threshold and can discover both positive and negative association rules. To capture the uncertainty associated with the fuzzy association rules, F-APACS uses a weight of evidence measure to represent confidence.

\item \emph{Kuok's Approach:}
Kuok et al. \cite{Kuok1998}, proposed the method for mining fuzzy association rules of the form, ``If \emph{X} is \emph{A} then \emph{Y} is \emph{B}.'' Here \emph{X}, \emph{Y} are attributes and \emph{A}, \emph{B} are fuzzy sets. This approach is important because it provides a better way of handling numeric attributes compared to existing methods. The study showed that the use of fuzzy sets helps to understand the correlation between two attributes through the significance factor and certainty factor.

\item \emph{Fuzzy Transaction Data mining Algorithm (FTDA):}
Hong et al. \cite{hong1999mining} used the fuzzy concept with the apriori algorithm to discover fuzzy association rules from a quantitative data set. To overcome the limitation of the apriori algorithm in handling quantitative data, the authors introduced the FTDA (Fuzzy Transaction Data mining Algorithm), which first transformed quantitative data into linguistic terms using membership functions. Next, the scalar cardinalities of all linguistic terms were calculated, and the apriori algorithm was modified to find association rules as fuzzy sets. However, a drawback of this method is that experts need to provide the best fuzzy sets of quantitative attributes manually.

\item \emph{Gyenesei's Approach:}
Gyenesei \cite{Gyenesei2001AFA} addressed the limitation of expert dependency in selecting fuzzy sets for quantitative attributes by introducing a fuzzy normalization process. To obtain unbiased membership functions, the author proposed using fuzzy covariance and fuzzy correlation values. Interest measures were defined in terms of fuzzy support, fuzzy confidence, and fuzzy correlation. The approach was evaluated using two methods: with normalization and without normalization. The non-normalized method produced the most interesting rules, while the number of rules generated by the normalized approach was comparable to the discrete method. The fuzzy normalization process helped to reduce anomalies that may arise from the arbitrary selection of fuzzy sets.

\item \emph{Generalized Fuzzy Quantitative Association Rule Mining Algorithm:}
Lee \cite{Lee2001} proposed a novel algorithm for generalized fuzzy QARM, incorporating fuzzy concept hierarchies for categorical attributes and fuzzy generalization hierarchies of linguistic terms for quantitative attributes. Unlike other methods, this approach calculates the weighted support and weighted confidence by taking into account the importance weights of attributes. To eliminate redundant rules, the $R$-interest measure is used. The algorithm converts each transaction into an augmented transaction and applies apriori \cite{agrawal1994fast} to generate frequent itemsets with the aid of weighted support and weighted confidence measures. It then extracts QARs by removing rules not meeting the R-interest measure's criteria.

\item \emph{Optimized Fuzzy Association Rule Mining(OFARM):}
Zheng et al. \cite{zheng2014optimized} proposed a novel algorithm, OFARM (optimized fuzzy association rule mining), in 2014 to optimize the partition points of fuzzy sets with multiple objective functions. The frequent itemsets are generated using a two-level iteration process, and the certainty factor with confidence is used to evaluate fuzzy association rules.
\end{itemize}

\paragraph{Hybrid Based Algorithms }
\begin{itemize}
\item \emph{Equal-Depth Partition with Fuzzy Terms (EDPFT):}
Zhang \cite{zhang1999mining} proposed an enhanced version of the \emph{equi-depth partition (EDP)} algorithm that integrated fuzzy terms, called EDPFT. This algorithm was designed to identify association rules that contain intervals, crisp values, and fuzzy terms on both the left-hand and the right-hand sides. Unlike FTDA, which relies on user-supplied fuzzy sets, EDPFT utilizes equi-depth partitioning to obtain the intervals of numeric attributes. Although the author did not evaluate the algorithm using any data set, this approach shows potential in dealing with both crisp and fuzzy values in ARM.

\item \emph{Mohamadlou et al. Algorithm:} 
Mohamadlou et al. \cite{Mohamadlou2009} introduced a fuzzy clustering-based algorithm for mining fuzzy association rules. The algorithm utilizes $C$-means clustering to cluster all the transactions, followed by obtaining the fuzzy partition for each attribute. It then converts the quantitative transactions into `fuzzy discrete transactions' by mapping the quantitative data into fuzzy partitions. The algorithm mines fuzzy association rules from the `fuzzy discrete transactions' using an ARM algorithm.

\item \emph{Fuzzy Inference Based on Quantitative Association Rule (FI-QAR):}
Wang et al. \cite{Wang2015} proposed a three-phase algorithm called FI-QAR, which integrates clustering and fuzzy techniques. In the first phase, the density-based fuzzy adaptive clustering (DFAC) \cite{DFAC} algorithm was applied to discretize numeric attributes into discrete intervals. The intervals were then combined with the TS fuzzy model to generate a nominal vector matrix, which was used to modify the A
apriori algorithm and reduce the scanning overhead of a large database. The second phase involved mining QARs using an improved apriori algorithm. Finally, the third phase pruned the association rules. The proposed approach offers a way to mine QARs from large databases effectively.

\item \emph{Fuzzy Class Association Rule Support Vector Machine (FCARSVM:)} Kianmehr et al. \cite{kianmehr2010fuzzy} proposed the FCARSVM to obtain fuzzy class association rules. The authors extracted Fuzzy Class Association Rules (FCAR) using a fuzzy $C$-means clustering algorithm in the first phase, and FCARs were weighted based on the scoring metric strategy in the second phase.
\end{itemize}

\subsubsection{The Optimization Method}

\paragraph{Evolution and DE-Based Algorithms}

\begin{itemize}
    \item \emph{GENetic Association Rules (GENAR):} Mata et al. \cite{mata2001mining} introduced GENAR as a genetic algorithm-based solution for NARM. GENAR is designed to identify numerical association rules with an unknown number of numeric attributes in the antecedent and a single attribute in the consequent. By utilizing genetic algorithms, GENAR offers an effective approach to discovering association rules involving numerical attributes.


\item \emph{Genetic Association Rules (GAR):}
The extended version of GENAR, called GAR, was proposed by Mata et al. \cite{mata2002discovering}. GAR utilizes the five fundamental phases of a genetic algorithm, namely initialization, evaluation, reproduction, crossover, and mutation, to discover intervals for numerical attributes. A key contribution of GAR is the introduction of a fitness function to determine the optimal amplitude for each numerical attribute's interval. The genes in GAR represent the upper and lower limits of the attribute intervals and are initially created randomly. Through crossover and mutation operations, a new generation of genes is generated, and the fitness function is used to evaluate the quality of the intervals. GAR provides an effective approach for identifying appropriate intervals for numerical attributes in ARM.


\item \emph{Genetic Association Rules Plus (GAR Plus):}
Alvarez et al. \cite{alvarez2012evolutionary} made enhancements to the GAR algorithm and introduced GAR Plus. This improved version enables the automatic extraction of intervals for numerical attributes through an evolutionary process, eliminating the need for pre-discretization. GAR Plus enhances the fitness function of GAR by incorporating additional parameters such as support, confidence, interval amplitude, and the number of attributes with a modifier. By considering these parameters, GAR Plus provides a more comprehensive evaluation of the fitness of candidate intervals, resulting in improved performance and accuracy compared to the original GAR algorithm.


\item \emph{Alatas and Akin Algorithm:}
Alatas and Akin \cite{alatacs2006efficient} have made significant contributions to the field of NARM. In one of their studies, Alatas extended the GAR algorithm to discover both positive and negative association rules. They compared the performance of their proposed algorithm with the original GAR algorithm and observed that the amplitude of the intervals generated by their approach was lower than that of GAR. This indicates that the extended algorithm by Alatas and Akin was able to identify more specific and precise intervals for numerical attributes, resulting in improved rule discovery.


\item \emph{QuantMiner:}
QuantMiner \cite{salleb2007quantminer} is a system for discovering QARs that employs a genetic algorithm. The system operates with a predefined set of rule templates, which can be either user-selected or computed by the system itself. These templates define the format of the QARs. By utilizing the genetic algorithm, QuantMiner searches for the optimal intervals for the numerical attributes specified in the rule templates. This approach allows the system to efficiently explore the search space and identify association rules that meet the desired criteria.


\item \emph{Expending Association Rule Mining with Genetic Algorithm (EARMGA):}
Yan et al. \cite{yan2009genetic} presented an encoding method for discovering association rules using a genetic algorithm. Their approach, named ARMGA, initially designed for boolean attributes, was extended to handle generalized association rules incorporating both categorical and quantitative attributes. The authors introduced a fitness function based on relative confidence, eliminating the need for a user-defined minimum support threshold. To handle quantitative attributes, they discretized them into intervals and integrated four genetic operators into the algorithm. The resulting enhanced version, EARMGA, successfully accommodated quantitative attributes and utilized the $k$-FP tree data structure for efficient rule mining.


\item \emph{Real-Coded Genetic Algorithm (RCGA):}
Martinez et al. \cite{martinez2010mining} introduced a real-coded genetic algorithm (RCGA) for NARM. The RCGA is a variation of the binary-coded CHC algorithm \cite{ESHELMAN1991}, known for its elitist selection mechanism that favors the best individual for the next generation. In the context of NARM, the RCGA is utilized to search for optimal intervals. By employing real-coded representations and incorporating the elitist selection feature, the RCGA aims to efficiently explore the search space and discover high-quality numerical association rules.


\item \emph{Quantitative Association Rules by Genetic Algorithm (QARGA):}
Martínez et al. \cite{martinez2011evolutionary} improved the RCGA by proposing QARGA to extract QARs from real-world multidimensional time series. The QARGA method discovered significant relationships between ozone concentrations in the atmosphere and other climatological time series, including temperature, humidity, wind direction, and speed.

\item \emph{Niching Genetic Algorithm for Quantitative Association Rules (NICGAR):} NICGAR was proposed by Martin et al. \cite{MARTIN2016} to prevent the generation of similar rules by reducing the set of quantitative rules, which includes positive and negative rules. The algorithm consists of three components: an external population, a punishment mechanism, and a restarting process to manage niches and avoid the same solutions. The article also proposes a new similarity measure to find the similarity between rules.

\item \emph{QAR-CIP-NSGA-II:} Martin et al. \cite{MARTIN2014} presented a novel multi-objective evolutionary algorithm called QAR-CIP-NSGA-II, which extends NSGA-II to simultaneously learn the intervals of attributes and conditions for each rule in a QAR system. QAR-CIP-NSGA-II aims to discover a set of high-quality QARs that balance interpretability and accuracy by maximizing comprehensibility, interestingness, and performance objectives. The algorithm incorporates an external population and a restarting method to enhance population diversity and store discovered nondominated rules. The comprehensibility of a rule is measured by the number of attributes involved in the rule, while the product of the certainty factor and support determines the accuracy. The interestingness measure, lift, is used to determine how significant the rule is.

\item \emph{Multi-Objective Genetic algorithm Association Rule mining (MOGAR):} Minaei-Bidgoli et al. \cite{Minaeibidgoli2013} proposed MOGAR algorithm for discovering association rules from numerical data.
The algorithm maintains a population of candidate association rules, representing potential solutions, and applies genetic operators such as selection, crossover, and mutation to evolve the population over successive generations. The fitness of each candidate rule is evaluated based on multiple objectives, such as confidence, interestingness and comprehensibility. MOGAR employs a Pareto dominance concept to identify non-dominated solutions
MOGAR has the ability to handle complex datasets with multiple conflicting objectives, providing a more comprehensive view of associations in the data.

\item \emph{Multi-Objective Positive Negative Association Rule Mining Algorithm (MOPNAR):} Martin et al. \cite{MARTIN2014mopnar} proposed MOPNAR, a multi-objective algorithm that aims to achieve the same objectives as QAR-CIP-NSGA-II, including mining a reduced set of positive and negative QARs. The authors also claimed to achieve a low computational cost and good scalability, even with an increased problem size. In addition, MOPNAR was compared with other existing evolutionary algorithms such as GAR, EARMGA, GENAR, and MODENAR.

\item \emph{Multi-Objective Quantitative Association Rule Mining (MOQAR):} Martínez et al. \cite{martinez2016improving} improved the multi-objective evolutionary algorithm (MOEA) non-dominated sorting genetic algorithm-II (NSGA-II) \cite{Deb2002} by integrating it with their proposed QARGA approach. The authors used principal component analysis (PCA) to select the best subset of quality measures for the fitness function. Additionally, different distance criteria were introduced to replace the crowding distance of solutions to obtain secondary rankings in Pareto fronts. The primary ranking was achieved through the non-dominated sorting of the solutions.

\item \emph{ Multi-Objective
Evolutionary Algorithm for Quantitative Association Rule Mining (MOEA-QAR):} The MOEA-QAR algorithm \cite{MOEA-QAR} combines a genetic algorithm with clustering to mine interesting association rules. The dataset is first clustered using $K$-means, and each cluster is used as input to a separate GA to extract rules for that cluster. The fitness function of each chromosome is defined by confidence, interestingness, and cosine2. The algorithm can be applied to the entire dataset or just to each cluster, and experiments show that more rules are retrieved per cluster than for the whole dataset. Notably, users do not need to specify minimum support or confidence thresholds.

\item \emph{Association Rule Mining with Differential Evolution (ARM-DE):} In 2018, Fister et al. \cite{Fister2018} proposed a novel approach to ARM with numerical and categorical attributes based on differential evolution. Their algorithm consists of three stages: domain analysis, solution representation, and fitness function definition. In domain analysis, attribute domains are determined for numerical and categorical attributes. For numerical attributes, the minimum and maximum bounds are defined, while for categorical attributes, a set of values is enumerated. Each solution is represented mathematically using a real-valued vector. The fitness function is then calculated based on confidence and support, and optimization is achieved by maximizing the fitness function value.

\item \emph{Rare-PEAR:} The Rare-PEARs algorithm proposed by Almasi et al. \cite{ALMASI2015} aims to discover various interesting and rare association rules by giving a chance to each rule with a different length and appearance. The algorithm decomposes the process of ARM into $N-1$ sub-problems, where each sub-problem is handled by an independent sub-process during Rare-PEARs execution. $N$ is the number of attributes, and each sub-process starts with a different initial population and explores the search space of its corresponding sub-problem to find rules with semi-optimal intervals for each attribute. This approach allows for a more comprehensive exploration of the search space, discovering more diverse and rare association rules.

\item \emph{Genetic Network Programming (GNP):} Taboada et al. \cite{taboada2008association} proposed Genetic Network Programming (GNP) as a graph-based approach to ARM with numerical attributes. GNP consists of three node types: a start node, a judgement node, and a processing node. The judgement nodes act as conditional branch decision functions, while the processing nodes act as action functions. Evolution is carried out using crossover and mutation operators, and the significance of important rules is measured using the chi-square test. Rules are stored in a pool, which is updated every generation, and the lower chi-squared value rule is exchanged with a higher chi-squared value rule. This approach effectively extracts important rules from the database. 
    
\item \emph{Grammar-Guided Genetic Programming Association Rule Mining (G3PARM):} Luna et al. \cite{luna2014reducing} applied Grammar-Guided Genetic Programming (G3P) to the task of finding QARs, building on their previous work in 2010, where they introduced G3PARM for ARM. The focus of the approach proposed in \cite{luna2014reducing} was to reduce gaps in numerical intervals and emphasize the distribution of instances. To achieve this, the authors developed a self-adaptive algorithm that dynamically adjusts the number of parameters used in the evolutionary process and utilizes context-free grammar to represent solutions. The algorithm aims to identify the best rules according to a given fitness function, which are then stored in a pool and updated in each generation.

\item \emph{Multi-Objective Differential Evolution algorithm for Numeric Association Rules (MODENAR):} Alatas et al. \cite{ALATAS2008modenar} proposed a multi-objective differential evolution algorithm to discover accurate association rules from numeric attributes. The algorithm was designed to optimize four objectives: amplitude, comprehensibility, support, and confidence, based on Pareto principles. The support and confidence of the discovered rules were required to be high. Comprehensibility was defined as the number of attributes involved in a rule, and shorter rules were preferred. The amplitudes of attribute intervals were aimed to satisfy fewer rules; hence amplitude was minimized while support, confidence, and comprehensibility were maximized. 

\end{itemize}

\paragraph{Swarm Intelligence-Based Algorithms}
\begin{itemize}
\item \emph{Rough Particle Swarm Optimization Algorithm (RPSOA):}

The RPSO algorithm was introduced as the first PSO-based algorithm for NARM with rough particles \cite{alatas2008rough}. This algorithm aims to determine numeric attribute intervals and then discover association rules that conform to these intervals, where the fitness function is responsible for determining the amplitude of the intervals. Rough values of each attribute are defined by upper and lower bounds and are useful in representing an interval for an attribute. Each rough particle has decision variables representing items and intervals. It consists of three parts: the first describes the antecedent or consequent of a rule, the second represents the lower bound, and the third represents the upper bound of the interval. An item is considered an antecedent if its value is between $0$ and $0.33$, a consequent if it is between $0.33$ and $0.66$, and if it is between $0.66$ and $1.0$, the item would not be included in the rule. Once the RPSO algorithm completes its execution, attribute bounds refinement is performed for the covered rule. This refinement step aims to improve the quality and accuracy of the discovered association rules by further optimizing the attribute bounds.


\item \emph{Chaotically ENcoded Particle Swarm Optimization Algorithm (CENPSOA):} 
The CENPSOA algorithm, proposed by Alatas and Akin \cite{ALATASCENPSO}, introduced the use of chaos variables and particles in PSO for the first time. Unlike previous PSO-based methods, CENPSOA employs chaotic numbers to encode particle information. Specifically, each chaotic number $mid_{rad}$ represents an interval with a lower bound of  $mid-rad$ and an upper bound of  $mid+rad$. In CENPSOA, a particle is represented as a string of chaotic parameters consisting of a midpoint and radius pair. Each decision variable consists of three parts: the first part represents the antecedent or consequent, the second part describes the midpoint and the third part represents the radius. This algorithm works similarly to RPSOA but is different only with the encoding of particles.

\item \emph{Parallel PSO for Quantitative Association Rule Mining (PPQAR):}
Yan et al.\cite{yan2019ppqar} parallelized the PSO algorithm for ARM to increase its scalability and efficiency in dealing with large datasets in real-world applications. To evaluate each particle's quality, the suggested technique used four optimization objectives: support, confidence, comprehensibility, and interest. The parallel PSO method employs two techniques to handle distinct application scenarios: particle-oriented and data-oriented. The particle-oriented technique is well-suited for small datasets with a large number of particles, treating each particle as a separate computing unit and computing the fitness function in parallel. On the other hand, the data-oriented approach is suitable for large datasets, dividing the entire dataset into partitions and treating each partition as a computing unit. Unlike the particle-oriented method, the data-oriented method updates particle locations, velocities, and local best sets in parallel. Both methods were compared with the benchmark serial algorithm.

\item \emph{Multi-Objective Particle swarm optimization
algorithm for Association Rules mining (MOPAR):}
The MOPAR algorithm, proposed by Beiranvand et al. \cite{beiranvand2014multi}, is a multi-objective particle swarm optimization (MOPSO) technique based on Pareto optimality. It aims to extract numerical association rules in a single step using three objectives: confidence, comprehensibility, and interestingness. Like RPSOA, the particle in MOPAR is represented by lower and upper bounds of intervals for each attribute. To address the problem of numerical ARM, MOPAR provides a redefinition of lbest and gbest particles and a selection procedure. The algorithm was compared with other multi-objective ARM algorithms, including MODENAR, MOGAR, RPSOA, and GAR.

\item \emph{PSO with the Cauchy Distribution (PARCD):}
A method proposed by Tahyudin et al. \cite{tahyudin2019improved} extends the MOPAR algorithm by combining PSO with the Cauchy distribution. In traditional PSO, the velocity of a particle approaches $0$ after many iterations, leading to premature searching and suboptimal results. The proposed approach addresses this issue by integrating the Cauchy distribution in the velocity equation, allowing particles to continue exploring the search space. This method uses multiple objectives, including support, confidence, comprehensibility, interestingness, and amplitude functions, to extract numerical association rules in a single step. To evaluate the method's performance, it was compared with MOPAR, MODENAR, MOGAR, and RPSOA on various datasets, and it was found that the proposed method, called PARCD, outperformed MOPAR.

\item \emph{Wolf Search Algorithm (WSA):}
Agbehadji \cite{wolf} introduced a wolf search algorithm for NARM inspired by the hunting behaviour of wolves. The algorithm is based on three stages of wolf-preying behaviour: actively seeking prey, passively seeking prey, and escaping from predators. The algorithm generates association rules if the wolf is actively seeking prey, and no rules are generated if the wolf is passively seeking prey or escaping. The fitness function includes support, confidence, the number of attributes, and the penalization of interval frequency. The algorithm represents rules using the wolf's best position and fitness value, and each wolf's position contains decision variables for items and intervals. While this study introduces the algorithm, it has not been evaluated on datasets, and the algorithm's accuracy and efficiency will be determined in future work.
 
\item \emph{Multi-objective Particle Swarm Optimization (MOPSO):}
The MOPSO algorithm, originally proposed by Coello \cite{Coello2004} in 2004, utilizes Pareto dominance and an archive controller. In 2019, Kuo et al. \cite{kuo2019multi} developed a MOPSO algorithm for NARM consisting of three stages: initialization, adaptive archive grid, and PSO searching. Particle representation and initialization are the same as in the RPSOA algorithm. The adaptive archive grid is a hypercube-shaped space designed to obtain non-dominated solutions by comparing all particle solutions using Pareto optimality. It contains two components: the archive controller and the grid. The external archive retains non-dominated solutions, and new solutions are added if existing ones do not dominate them or if the external archive is empty, the new solution is saved in the external archive; otherwise, it is discarded. The adaptive grid approach is used when the external population reaches its maximum capacity. The objective function space is partitioned into regions. The grid is recalculated if the external population's individual falls outside the grid's bounds, and each individual within it must be relocated. After the archive grid stage, PSO searching occurs. The algorithm also utilizes three objectives: confidence, comprehensibility, and interestingness to generate rules.

\item \emph{Ant Colony Optimization for Continous attributes ($ACO_R$):}
The $ACO_R$ algorithm, introduced by Moslehi and Eftekhari \cite{moslehi2011multi}, is an ant colony optimization technique designed to discover association rules for numeric attributes without relying on minimum support and confidence thresholds. Unlike the ACO algorithm, which uses a discrete probability distribution, $ACO_R$ employs a probability density function. It employs a solution archive size of $k$ to describe the pheromone distribution over the search space, instead of a pheromone table. The algorithm works by having the ants move across the archive, selecting a row based on its associated weight $(\omega)$. Then a new solution is created by sampling the Gaussian function $g$ for each dimension's values in the selected solution. Each numeric attribute corresponds to one dimension of the solution archive, which is divided into three sections that make up a numeric association rule: the first part represents the rule's antecedent or consequence; the second part represents its value; and the third part represents its standard deviation, which is used to form numeric attribute intervals.

The algorithm uses Gaussian functions to determine the attribute intervals that correspond to interesting rules, with the function controlling the intervals' frequency and length. The objective function has four components. The first section, which can be viewed as the rule's support, measures the importance of the association rule. The second section is the confidence value of the rule. The third section is the number of attributes, while the last section penalizes the amplitude of the intervals that comply with the itemset and rules. The pheromone update technique introduces a set of new solutions, each generated by one ant, and eliminates the same number of bad solutions from the archive after ranking them to track the solutions. This ensures that the top-ranked solutions are always at the archive's top, and that the best solution in each execution of $ACO_R$ is a rule.

\item \emph{Multi-Objective Cuckoo search Algorithm for Numerical Association Rule Mining (MOCANAR):}
MOCANAR \cite{kahvazadeh2015mocanar} is a multi-objective cuckoo search algorithm that uses Pareto principles to derive high-quality association rules from numeric attributes. The algorithm mimics the brooding parasitic behavior of cuckoo species and represents ARM using a $2D$ array. The columns of the array represent the attributes in the dataset, and the first row among three rows represents the attribute's location. The second row consists of the lower bound of the attribute, and the third row represents the upper bound of the attribute. A value of $0$ in the first row indicates that the related attribute is not present in the rule, $1$ shows that the attribute belongs to the antecedent part of the rule, and $2$ shows that the attribute belongs to the consequent part of the rule.
MOCANAR considers four objectives: support, confidence, interest, and comprehensibility. The algorithm was evaluated on three datasets and produced a small number of high-quality rules incrementally for each iteration of the method.

\item \emph{Multi-Objective Bat Algorithm for Numerical Association Rule Mining (MOB-ARM):}
Heraguemi et al. \cite{heraguemi2018mBAT} proposed a multi-objective bat algorithm for NARM inspired by microbats' behaviour. The algorithm uses four quality measures, namely support, confidence, comprehensibility, and interestingness, and two global objective functions to extract interesting rules. The first objective function combines support and confidence, while the second objective function considers comprehensibility and interestingness. The algorithm comprises three main steps: initialization, searching for the non-dominance solution for the Pareto point, and searching for the best solution for each bat at the Pareto point. The rule is encoded using the Michigan approach. The bats are initialized with random frequency and velocity, and the proposed algorithm is also compared with other algorithms, including MODENAR, MOGAR, and MOPAR.

\item \emph{Discrete Crow Search Algorithm for Quantitative Association Rule Mining (DCSA-QAR):}
In 2021, a new algorithm called DCSA-QAR was proposed for mining numerical association rules \cite{ledmi2021discrete}. This approach utilizes a novel discretization algorithm called Confidence-based Unsupervised Discretization Algorithm (CUDA) that employs the confidence measure to discretize numerical attributes. The CSA is then transformed from continuous to discrete using crow position encoding, and new operators are used to ensure that any position update within the search space is valid. Each crow in the flock is represented by its current position and memory positions, with each particle composed of two vectors for control and parametric attributes. The control attributes can have one of three values: $0$ indicates that the attribute is not part of the rule, $1$ indicates that it belongs to the antecedent, and $-1$ indicates that it is part of the consequent. The fitness function is optimized by maximizing the measures of support, confidence, and gain of the rules. DCSA-QAR was compared with several mono and multi-objective algorithms, including NICGAR, MOPNAR, MODENAR, and MOEA-Ghosh.
\end{itemize}
\paragraph{Physics Based Algorithms}
\begin{itemize}
\item \emph{Gravitational Search Algorithm for NARM (GSA-NARM):}
GSA is a physics-inspired metaheuristic that leverages Newton's law of gravity. In the context of NARM, the GSA algorithm, as described by Can et al. \cite{can2017automatic}, aims to discover attribute intervals simultaneously without needing a minimum support or confidence threshold. In GSA-NARM, agents are treated as objects, and their positions represent potential solutions. The objective function determines the amplitude of the intervals being explored. The algorithm identifies the position of the agent with the heaviest mass as the global solution, analogous to the gravitational force exerted by massive objects in Newton's law. During the optimization process, the fitness function is evaluated for each agent, and the gravitational constant, denoted as $G$, is updated based on the performance of the best and worst agents in the population. The mass $M$ of each agent is computed, and the velocity and position are updated accordingly, mimicking the motion of celestial objects influenced by gravitational forces. The GSA-NARM algorithm continues iterating until the stopping criteria are met, such as reaching a maximum number of iterations or achieving a desired fitness value. The algorithm then returns the association rule with the best fitness value obtained during the optimization process. GSA-NARM has demonstrated promising results when compared to other state-of-the-art methods for NARM, showcasing its effectiveness in tackling the NARM problem.


\end{itemize}

\paragraph{Algorithm for Hybrid Based}
\begin{itemize}
\item \emph{Hybrid Genetic PSO-Quantitative Association Rule Mining (HGP-QAR):}
Moleshi et al. \cite{moslehi2020novel} introduced a hybrid approach called HGP-QAR, which combines the strengths of multi-objective GA and multi-objective PSO methods. By leveraging the advantages of both techniques, HGP-QAR aims to improve the efficiency of NARM. 
The hybridization of GA and PSO allows for the exploration of the search space from different perspectives. In HGP-QAR, individuals are represented as chromosomes for GA and particles for PSO. The individuals are sorted based on a fitness function that considers three metrics: confidence, interestingness, and comprehensibility. During the optimization process, the upper half of individuals follow the stages of GA, including selection, crossover, and mutation, while the lower half follows the stages of PSO, updating their velocity and positions based on the personal best (pbest) and global best (gbest) positions. This combination of GA and PSO allows for a more efficient search and exploration of the solution space. The outcomes obtained from GA and PSO are then combined to generate the next generation and form new rules. This process is repeated until the termination criteria are met, such as reaching a maximum number of iterations or achieving satisfactory results. Various experimental results show that the hybrid GA-PSO approach, HGP-QAR, outperforms other algorithms like MOPAR and PARCD in terms of efficiency, demonstrating its effectiveness in NARM.

\item \emph{Multi-objective Hybrid Differential Evolution Sine Cosine Numerical Association Rule Mining Algorithm (MOHDESCNAR):} DE has been known to suffer from premature convergence and stagnation issues in multi-modal search spaces. A recent approach called MOHDESCNAR \cite{ALTAY2021DEsine} has been proposed to overcome these problems. This algorithm reduces the number of numerical association rules by adjusting the intervals of related numeric attributes. It employs hybrid sine and cosine operators with DE, which can overcome stagnation issues. The proposed algorithm balances exploration and exploitation by using global DE exploration and local SCA exploitation during iterations to prevent premature convergence and stagnation problems. This study used three methods: using only the sine operator (MOHDESNAR), only the cosine operator(MOHDECNAR), and both the sine and cosine operators (MOHDESCNAR).

\item \emph{Quantitative Association Rule miner with Chaotically Encoded Hybrid Differential Evolution and Sine Cosine Algorithm (QARCEHDESCA):}
Altay and Alatas proposed the MOHDESCNAR algorithm in 2021, which used a combination of DE and the sine and cosine algorithms. In 2022, the authors introduced a new hybrid algorithm called QARCEHDESCA \cite{altay2022chaos}, which employs chaos number-based encoding and HDESCA (Hybrid differential evolution sine cosine algorithm). The QARCEHDESCA algorithm dynamically discovers the ranges of quantitative attributes and association rules. It randomly initializes candidate search agents to find quantitative associations. The initial set of search agents is removed from all-dominating search agents. The remaining nondominated search agents are sent to SCA-based new operators and DE crossover. The nearest neighbour distance function is used to remove rules close to each other when the count of nondominated rules exceeds the defined threshold. For QARCEHDESCA, the best search agent and one random agent are chosen for sine and cosine operators. After that, DE's crossover operator is applied to nondominated search agents. If the trial agents dominate the target search agent, it is added to the population; otherwise, the search agent with the highest weighted sum fitness is chosen for subsequent iterations. When the maximum number of iterations is reached, QARCEHDESCA returns nondominated QARs. The fitness function of the algorithm aims to maximize support, confidence, and comprehensibility while minimizing attribute amplitudes. Each search agent represents a numerical association rule with two components: inclusion/exclusion and a chaotic number representing the center point and radius. QARCEHDESCA is compared with RPSOA and other intelligent optimized algorithms.

\end{itemize}

\subsubsection{The Statistical Method}
\begin{itemize}
 \item \emph{Aumann and Lindell's Work:}
Aumann and Lindell \cite{aumann2003statistical} introduced a new definition of QARs based on the distribution of values of quantitative attributes and presented an algorithm to mine them. To consider the distribution of continuous data, they used conventional statistical measures.

\item \emph{Webb's Work:} 
Aumann and Lindell's approach has the disadvantage of being impractical for generating frequent itemsets in dense data. To address this limitation, Webb proposed an efficient admissible unordered search algorithm for discovering impact rules, which capture meaningful interactions between data selectors and numeric variables in dense data \cite{webb2001discovering}. Impact rules were introduced as a new name for QARs. The proposed OPUS\_IR algorithm uses the OPUS framework and does not need to retain all frequent itemsets in memory during frequent itemset generation, unlike Aumann and Lindell's method \cite{aumann2003statistical}. It also does not require a minimum cover to be specified for the search. The OPUS\_IR algorithm was compared with the frequent itemset approach in terms of performance.

\item \emph{Kang et al. Work:} 
The authors of the study \cite{kang2009bipartition} introduced a new approach to bipartition quantitative attributes called \emph{standard deviation minimization}. This technique minimizes the standard deviation of two partitions obtained by dividing the attribute into two parts, and it outperforms existing bipartition techniques. The authors also redefined the mean-based and median-based bipartition techniques, and their experimental results confirmed the effectiveness of the proposed framework.
\end{itemize}

\subsubsection{Miscellaneous Other Methods}

\begin{itemize}
\item \emph{Mutual Information and Clique (MIC) Framework:}
Yiping et al. \cite{ke2008information} proposed a novel approach for mining QARs using an information-theoretic framework called MIC. This framework avoids the generation of excessive itemsets by investigating the relationship between attributes. The approach comprises three phases: 1) discretization, which partitions numeric attributes into intervals; 2) MI graph construction, which computes the normalized mutual information of attributes and represents their strong relationships using a MI (mutual information) graph; and 3) clique computation and QAR generation, which computes frequent itemsets using cliques and generates QARs. The experiments demonstrate the effectiveness of the MIC framework in reducing the number of generated itemsets and improving the efficiency of QAR mining.

\item \emph{Generalized One-sided Quantitative Association Rule mining (GOQR) and Non-redundant Generalized One-sided Quantitative Association Rule mining (NGOQR):} 
Zhiyanget al.\cite{hu2022cognitive} proposed a cognitive computing-based method for NARM, consisting of two algorithms: \emph{GOQR} and \emph{NGOQR}. These algorithms consider the order relation of attribute values when mining rules. The first phase of the \emph{GOQR} algorithm generates frequent itemsets, while the second phase extracts generalized one-sided QARs. To enhance efficiency, the rules are reduced using a generalized one-sided concept lattice. For non-redundant rule extraction, the \emph{NGOQR} algorithm first executes the minimal generator of a target itemset algorithm and then continues with rule mining.

\item \emph{Quantitative Miner with the VMO algorithm (QM\_VMO):}
The Quantitative Miner with the VMO algorithm \emph{(QM\_VMO)} \cite{QMVMO} utilizes the Variable Mesh Optimization algorithm \cite{VMO2012}, a population-based meta-heuristic. The algorithm represents the population $P$ as a mesh of $n$ nodes $P = n_1, n_2,..., n_n$, where each node corresponds to a possible solution and consists of an m-dimensional vector $n_i = (v^i_1, v^i_2,..., v^i_m)$. The algorithm primarily operates through expansion and contraction processes. \emph{QM\_VMO} is executed in three stages: (i) defining a rule template, (ii) generating the rule population, and (iii) optimizing the numerical attributes of the rule by optimizing intervals.
Compared with QuantMiner, \emph{QM\_VMO} is found to be less sensitive to changes in the dataset.
\end{itemize}

\subsection{RQ3 What are the advantages and limitations of the existing NARM methods?}
There are strengths and limitations associated with each method for NARM. These advantages and limitations of each approach are summarized in Tables \ref{advantagedis}, \ref{advantageop}, and \ref{advantagest}. 


 Discretization methods are advantageous in terms of simplicity, interpretability, and flexibility. They allow for the handling of both categorical and numerical attributes, and the resulting discrete intervals can be easily understood and applied. However, these methods require the specification of a user-defined threshold, which can be subjective and may affect the quality of the discovered rules. Discretization can also lead to information loss and may not capture the true underlying patterns in the data. 

 Optimization methods excel in their ability to discover relationships and patterns in high-dimensional data without the need for user-defined thresholds or discretization steps. They can handle both categorical and numerical attributes and are often more robust to noise and missing data. However, these methods can suffer from issues such as convergence problems, finding only local optima, high computational complexity, and the need for a large amount of computational resources.

 Statistical methods are advantageous in their ability to handle missing data and noise. They are also well-suited for analyzing categorical data and can provide statistical significance measures for the discovered rules. However, these methods are typically designed for categorical data and may not be suitable for numerical attributes. They often assume linear relationships and may not capture more complex patterns present in the data.
Overall, each approach has its strengths and limitations, and the choice of method depends on the specific requirements and characteristics of the dataset being analyzed.

\begin{longtable}{p{1.5cm} p{1.2cm} p{4.9cm} p{4.9cm}}
    \caption{Advantages and Limitations of Discretization Method}
  \\
    \hline
    Approaches & Reference & Advantages & Limitations\\ \hline
    \endhead

 Partitioning & \cite{srikant1996} & Simple and easy to implement. & Adjusting minimum support and minimum confidence. \\ 
 & \cite{chan1997effective}& Discover both +ve and -ve rules.  Avoid user-specified threshold. & need to adjusted difference analysis \\
 \hline
Clustering & \cite{lent1997clustering} & The ARCS system scales better than linearly with data size. & Algorithm is sensitive to noise.
 Not used for high dimensional data.\\
& \cite{wang1998interestingness} & Scalable for very large databases. Generate only non-overlapping intervals. & Only generate the rules where consequent should be categorical attribute \\
& \cite{mg2000} & Reflects all the possible interdependencies between attributes in data sets  & Requirement of the user-specified threshold. \\
& \cite{lian2005efficient}&  Efficiently identify a small set of subspaces for finding dense regions.
In each cover searching is limited.
 & Need to specify many thresholds.
 Performance is poor on more than 10 dimensions. The dimensionality curse problem is unsolved. The algorithm does not perform well for the data set with uniform density. \\
  & \cite{Guo2008} & Effective and scale up linearly with an increased number of attributes. &  Minimum threshold is needed. \\ 
  & \cite{Yang2010} & There is no need to scan the database many times. Do not generate many candidate units. Histogram H' saves the calculation time of support of each grid. & As the number of transactions increases, the run time also increases.\\
  \hline

 \hline
 
Fuzzy &\cite{zhang1999mining} & Prune less interesting rules.&  \\
& \cite{hong1999mining} &  Accuracy increased as the number of transactions increased. & Membership functions should be known in advance. \\
& \cite{zheng2014optimized} & Optimized the fuzzy sets. The frequent itemsets are created through a two-level iteration process. Flexible membership function.  & \\

& \cite{Wang2015} &  Provide better clustering than other methods. & \\
\hline
   \label{advantagedis}
\end{longtable}

\begin{longtable}{p{1.5cm} p{1.3cm} p{5.5cm} p{4.2cm}}
    \caption{Advantages and Limitations of Optimization Method}\\
    \hline
    Approaches & Reference & Advantages & Limitations\\ \hline    
    \endhead
Evolution and DE &\cite{mata2001mining} & Find association rules from numeric dataset without discretization &\\
& \cite{mata2002discovering} & Find the amplitude of the intervals by fitness function. & Only frequent itemsets are generated.\\
& \cite{alvarez2012evolutionary} & Find association rules from numeric and categorical without discretization & \\
& \cite{alatacs2006efficient} & discover both positive and negative rules. &  \\
& \cite{yan2009genetic} & High-performance
association rule mining, System automation, no need for user-specified minimum support threshold & \\
 & \cite{MARTIN2016}  & Low run time, discover diverse, both positive and negative rules.&  \\
& \cite{martinez2016improving} & Low computational cost  and good scalability & \\
& \cite{MOEA-QAR} & No need to determine the minimum support and minimum confidence. & \\
& \cite{Fister2018} & Capable of dealing with numerical and categorical attributes. & The algorithm is unable to shrink the lower and upper borders of the numerical attributes. \\
& \cite{ALATAS2008modenar} & association rules are mined without generating frequent itemsets. The algorithm is easy to implement and independent from the requirement of minimum support and minimum confidence threshold. &  DE suffers from stagnation and premature convergence problem and its local exploitation capability is weak.\\
\hline
Swarm-Intelligence &\cite{yan2019ppqar} & Efficient and scalable to process huge dataset. & PSO trap in local optima.\\
& \cite{beiranvand2014multi} & Prevent generation of huge useles rules. No requirement for  minimum support and minimum confidence threshold. & Low support values for association rules.\\
& \cite{tahyudin2019improved} & Increase the global optimal value of expanded search space. & \\
& \cite{moslehi2011multi} & No need for minimum support and minimum confidence threshold. & Variable correlations are not differentiated by ant algorithms.\\
& \cite{kahvazadeh2015mocanar} & Provide better support and confidence. & Higher number of extracted rules decreases the interpretability of the results.\\
& \cite{heraguemi2018mBAT} & Reduces computation time. & \\
\hline
Physics-based &\cite{can2017automatic} & The confidence and support values of the automatically mined rules are very high. No prior requirement for minimum support and confidence threshold. The problem of attribute interactions has been solved. & Not very efficient in searching.\\
\hline
Hybrid &\cite{altay2022chaos} & Efficient with respect to the mean number of rules, mean confidence,
and mean size metrics.& Does not provide the higher mean support value.\\
\hline
    \label{advantageop}
\end{longtable}

\begin{table}[!ht]
    \centering
    \caption{Advantages and Limitations of Statistical Method}
    \label{advantagest}
    \begin{tabular} {p{1.4cm} p{5.5cm} p{5.5cm}}
    \hline
    Reference & Advantages & Limitations\\ \hline
\cite{aumann2003statistical} & Able to handle large datasets. Adaptable and may be used for various types of data and user needs. & Computationally expensive. Assump\-tions-dependent and limited interpretability.\\
\cite{webb2001discovering} & Its unordered search quality makes it handle different types of data. High performance. Identify high-impact patterns and relationships. & High computation cost for large datasets. Limited interpretability.\\
 
\cite{kang2009bipartition} & Accuracy of ARM can be improved by identifying patterns and relationships within particular subsets of the data. Able to handle large datasets. & Difficult to choose the right threshold. Loss of information could be possible.\\
\hline
\end{tabular}
\end{table}

\subsection{RQ4 Which objectives are considered by the several existing multi-objective optimization NARM algorithms?}
Optimization problems are prevalent and important in scientific research. They can be categorized into two types based on the number of objective functions: single-objective and multi-objective optimization problems. In NARM, the most commonly used parameters are support and confidence, making NARM algorithms single-objective optimization methods where a single solution is selected based on the user's requirements. On the other hand, multi-objective optimization problems involve computing multiple objective functions simultaneously, which can conflict with each other.  A solution that works well for one function may be ineffective for another. This makes finding a single solution that satisfies all objectives difficult, and instead, a set of Pareto-optimal solutions is obtained that trade-off between the competing objectives. Table \ref{multiobjective} lists the objectives considered in multi-objective optimization NARM studies, and Table \ref{multiobjective algo} provides the names of algorithms that utilize these objectives. A detailed explanation of all these objectives is given via Eq. \ref{eq5}--\ref{eq13}.

\begin{table}
    \centering
    \caption{List of Objectives for Multi-objective Optimization Algorithm for NARM}
    \label{multiobjective}
    \begin{tabular} {p{2.5cm} p{10cm}}
    \hline
    Objectives & References \\ \hline
Confidence &\cite{martinez2016improving,MOEA-QAR,ALTAY2021DEsine,ALATAS2008modenar,yan2019ppqar,beiranvand2014multi,tahyudin2019improved,kuo2019multi,moslehi2011multi,Minaeibidgoli2013,kahvazadeh2015mocanar,heraguemi2018mBAT,ledmi2021discrete,altay2022chaos,moslehi2020novel,ALMASI2015}\\
Support &\cite{ALTAY2021DEsine,ALATAS2008modenar,yan2019ppqar,tahyudin2019improved,moslehi2011multi,kahvazadeh2015mocanar,heraguemi2018mBAT,ledmi2021discrete,altay2022chaos} \\
Interestingness &\cite{MARTIN2014,MARTIN2014mopnar,MOEA-QAR,yan2019ppqar,beiranvand2014multi,tahyudin2019improved,kuo2019multi,moslehi2011multi,Minaeibidgoli2013,kahvazadeh2015mocanar,heraguemi2018mBAT,moslehi2020novel,ALMASI2015} \\
Comprehensibility &\cite{MARTIN2014,MARTIN2014mopnar,ALTAY2021DEsine,ALATAS2008modenar,Minaeibidgoli2013,yan2019ppqar,beiranvand2014multi,tahyudin2019improved,kuo2019multi,kahvazadeh2015mocanar,heraguemi2018mBAT,altay2022chaos,moslehi2020novel}  \\
Amplitude &\cite{ALTAY2021DEsine,ALATAS2008modenar,tahyudin2019improved,moslehi2011multi,altay2022chaos} \\
Performance &\cite{MARTIN2014,MARTIN2014mopnar} \\
Accuracy  &\cite{martinez2016improving,ALMASI2015} \\
Leverage &\cite{martinez2016improving} \\
Gain  &\cite{ledmi2021discrete} \\
Cosine  &\cite{MOEA-QAR} \\
\hline
\end{tabular}
\end{table}

\begin{table}
    \centering
    \caption{ List of Multi-Objective Algorithms for NARM}
    \label{multiobjective algo}
    \begin{tabular} {p{3cm} p{1.5cm} p{7.5cm}}
    \hline
    Algorithms & Reference & Objectives \\ \hline

MOGAR &\cite{Minaeibidgoli2013} &
Confidence, Comprehensibility, Interestingness \\
QAR-CIP-NSGA-II &\cite{MARTIN2014} & Comprehensibility, Interestingness, Performance \\

MOPNAR &\cite{MARTIN2014mopnar} & Comprehensibility, Interestingness, Performance \\
MOQAR &\cite{martinez2016improving} & Accuracy, Leverage, Confidence \\
MOEA-QAR &\cite{MOEA-QAR} & Confidence, Interestingness, Cosine\\
Rare-PEAR &\cite{ALMASI2015} & Interestingness, Accuracy, Confidence\\
MODENAR &\cite{ALATAS2008modenar} & Support, Comprehensibility, Confidence, Amplitude\\
MOHDESCNAR &\cite{ALTAY2021DEsine} & Support, Comprehensibility, Confidence, Amplitude\\
PPQAR &\cite{yan2019ppqar} & Support,
Confidence, Comprehensibility, Interestingness\\
MOPAR &\cite{beiranvand2014multi} & Confidence, Comprehensibility, Interestingness\\
PARCD &\cite{tahyudin2019improved} &  Support, Confidence, Comprehensibility, Interestingness, Amplitude\\
MOPSO &\cite{kuo2019multi} & Confidence, Comprehensibility, Interestingness\\
$ACO_R$ &\cite{moslehi2011multi} & Support, Confidence, Interestingness, Amplitude\\
MOCANAR &\cite{kahvazadeh2015mocanar} &  Support, Confidence, Interestingness, Amplitude\\
MOB-ARM &\cite{heraguemi2018mBAT} & Support, Confidence, Comprehensibility, Interestingness\\
DCSA-QAR &\cite{ledmi2021discrete} & Support, Confidence, Gain\\
QARCEHDESCA &\cite{altay2022chaos} & Support, Comprehensibility, Confidence, Amplitude\\
HGP-QAR &\cite{moslehi2020novel} & Confidence, Comprehensibility, Interestingness\\
\hline
\end{tabular}
\end{table}

\paragraph{Support:}
The number of records with both $X$ and $Y$ itemsets determines the rule's support count. |D| is the total number of records in a dataset.

\begin{equation}
\label{eq5}
 \textit{Support} (X \Rightarrow Y) = \frac{|(X\cup Y)|}{|D|}
\end{equation}

\paragraph{Confidence:}
The confidence metric assesses the quality of a rule by counting the number of times an AR appears in the entire dataset. 
The following equation \ref{eq6} is used to compute the confidence of the rule $X \Rightarrow Y$. Furthermore, these parameters do not ensure that significant rules will be generated.

\begin{equation}
\label{eq6}
 \textit{Confidence}(X \Rightarrow Y) = \frac{\lvert(X\cup Y)\rvert}{\lvert X \rvert}
\end{equation}

\paragraph{Interestingness:}
The interestingness of a rule is a metric for determining how surprising a rule is to users, not just all possible rules. The first component of Eq.(\ref{eq7}) relates to the probability of producing the rule based on the antecedent part, the second part relates to the probability of producing rules based on the consequent part, and the last component is the probability of producing rules based on the overall dataset.

\begin{equation}
\label{eq7}
\textit{Interestingness} =  \frac{\textit{Support}|(X\cup Y)|}{\textit{Support}|X|} 
\cdot \frac{\textit{Support}|(X\cup Y)|}{\textit{Support}|Y|} \cdot 
\Biggl (
{1-\frac{\textit{Support}(X\cup Y)}{\textit{Support}(X)}}
\Biggr )
\end{equation}

\paragraph{Comprehensibility:}
The number of attributes included in both the antecedent and consequent parts of the rule is measured by comprehensibility \cite{Ghosh2004}. If the generated rules contain more attributes, then the rules will be difficult to comprehend. The rule is more comprehensible if the number of conditions in the antecedent part is less than that in the consequent part. The following expression measures the comprehensibility of an association rule: 

\begin{equation}
\label{eq8}
\textit{Comprehensibility} = \frac{\log(1+ |Y|)}{\log(1+ |X\cup Y|)}
\end{equation}
Where $|Y|$ and $|X\cup Y|)$ represent the number of attributes in the consequent part and both parts.

\paragraph{Amplitude:}
The intervals in each attribute that comply with interesting rules must have smaller amplitudes. If two rules have the same number of rows and attributes, the one with smaller intervals will provide more information. Amplitude is a minimization function; however support,
confidence and comprehensibility are maximization functions \cite{ALATAS2008modenar}. 

\begin{equation}
\label{eq9}
Amplitude \: of \, the \: Intervals = 1- \frac{1}{m} \sum_{i=1}^{m} \frac{u_i - l_i}{max(A_i) - min(A_i)}
\end{equation}

\paragraph{Performance:}
The product of support and CF is performance. Performance enables the ability to mine accurate rules with a suitable trade-off between local and general rules.
This measure has a range of values between $0$ and $1$.
The user may find a rule with a performance value close to $1$ more useful.

\paragraph{Accuracy:}
Accuracy represents the veracity of the rule \cite{Hamilton2006}.

\begin{equation}
\label{eq10}
 \textit{Accuracy}(X\Rightarrow Y) = \textit{Support} (X\Rightarrow Y) + \textit{Support} (\neg X \Rightarrow \neg Y)
\end{equation}

\paragraph{Leverage:}
Leverage is the difference between the frequency with which the antecedent and the consequent are identified together and the frequency with which they would be expected to be observed together, given their individual support \cite{piatetsky1991}. It represents the strength of the rule.

\begin{equation}
\label{eq11}
    \textit{Leverage}(X\Rightarrow Y) = {\textit{Support}(X\cup Y)} - {\textit{Support}(X)}\cdot{\textit{Support}(Y)}
\end{equation}

\paragraph{Gain:}
Gain is the difference between the confidence of both the antecedent and consequent part and the support of the consequent part \cite{Hamilton2006}.

\begin{equation}
\label{eq12}
    \textit{Gain}(X\Rightarrow Y) = {\textit{confidence}(X\Rightarrow Y)} - {\textit{Support}(Y)}
\end{equation}

\paragraph{Cosine:}
The cosine measure considers both the pattern's interest and its significance \cite{tan2004selecting}.

\begin{equation}
\label{eq13}
  \textit{Cosine}(X\Rightarrow Y) = \frac{\textit{Support}(X\cup Y)}{\sqrt{{\textit{Support}(X)}\cdot{\textit{Support}(Y)}}}
\end{equation}


\subsection{RQ5 What are the metrics to evaluate the NARM algorithms?}

\begin{figure}
  \centering
  \includegraphics[width=\linewidth]{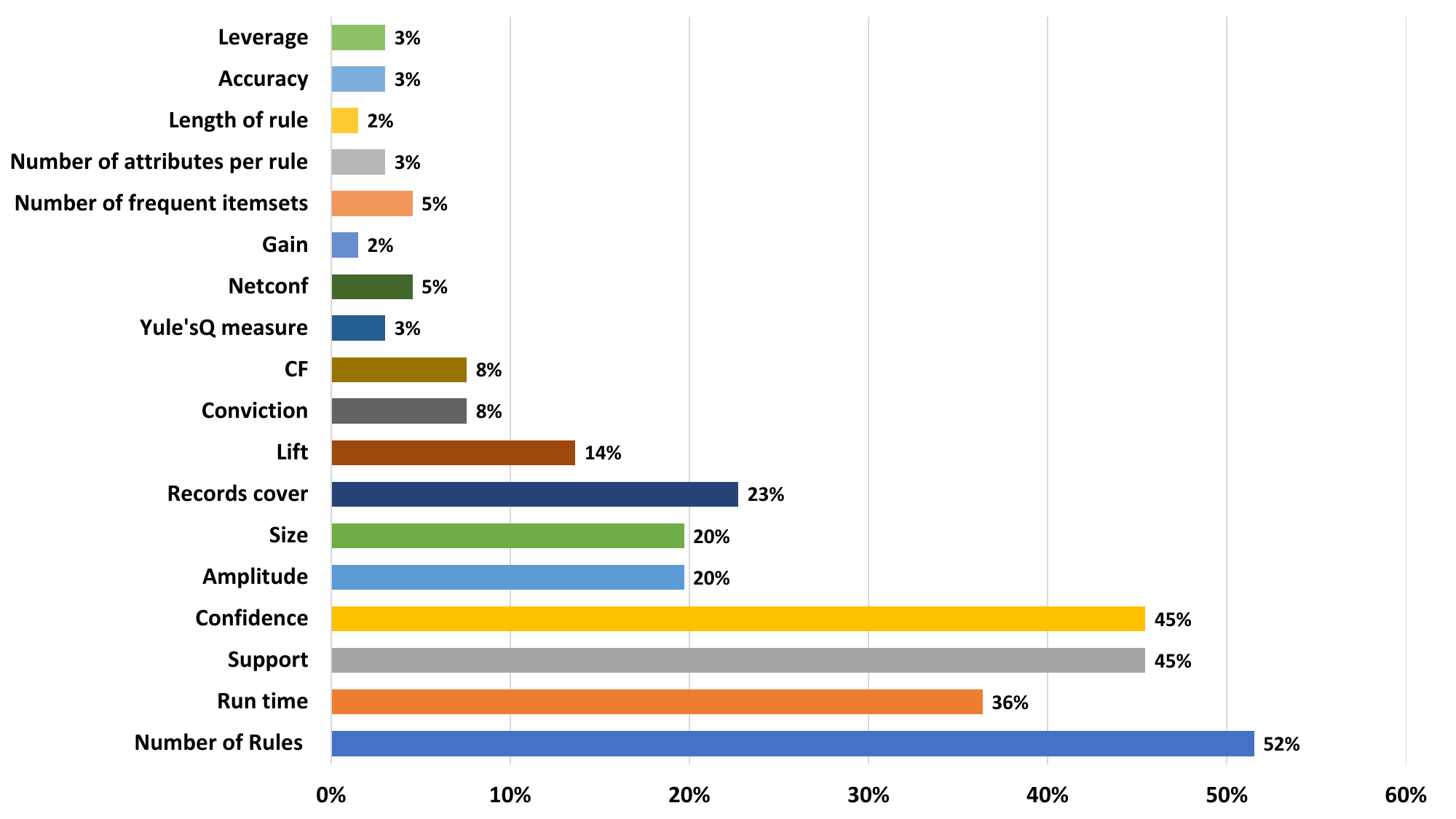}
  \caption{Metrics Used to Evaluate NARM Algorithms.}
  \label{metrics}
\end{figure}

\begin{figure}
  \centering
  \includegraphics[width=\linewidth]{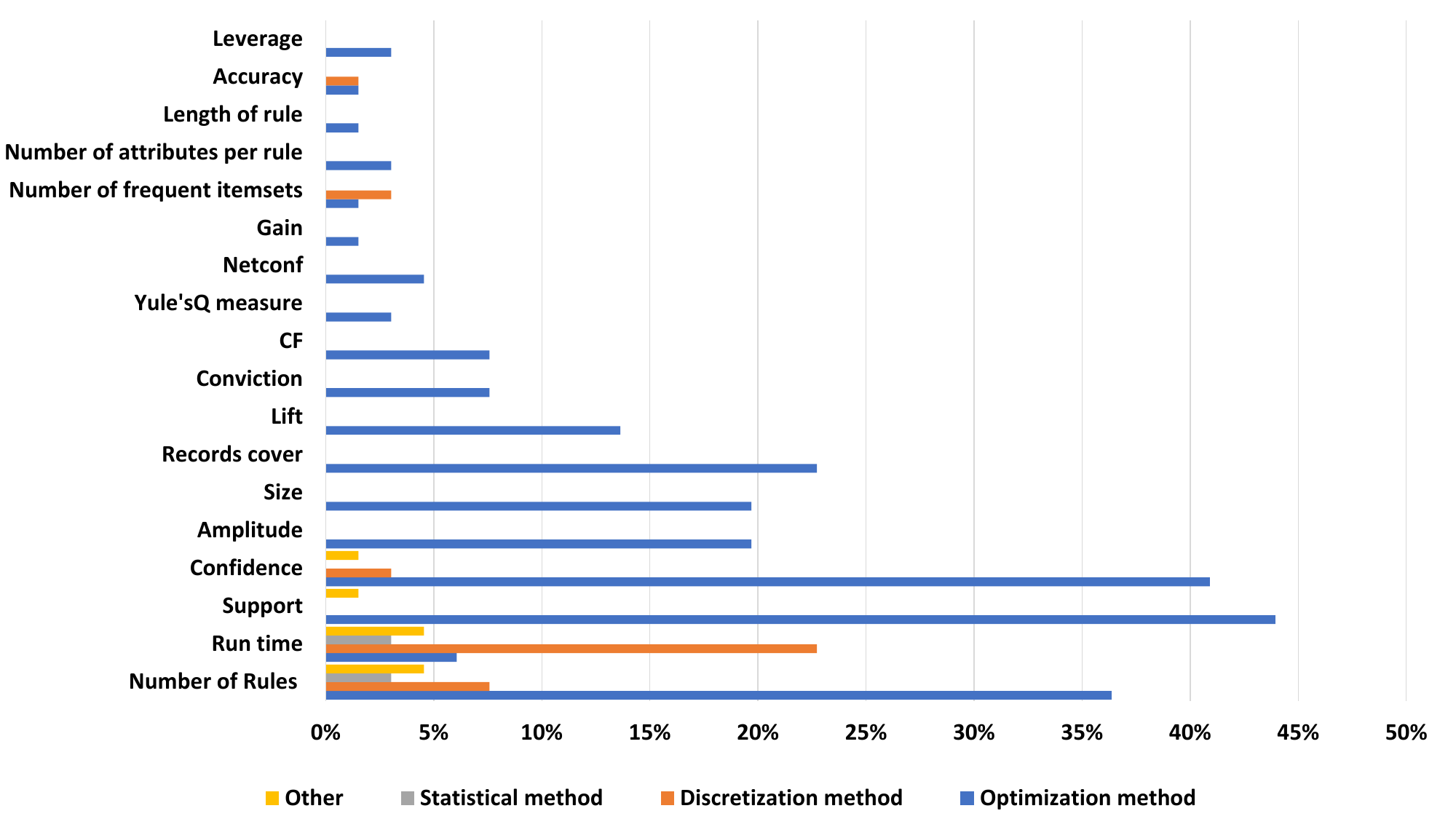}
  \caption{ Distribution of Metrics Used in NARM Methods.}
  \label{metricsdis}
\end{figure}

This RQ aims to identify the commonly used evaluation metrics in NARM algorithms.  As shown in Figure \ref{metrics}, the number of rules is the most commonly used metric, followed by support, confidence, and run time. Only a few algorithms use other metrics, such as Yule'sQ measure ($3$\%), leverage ($3$\%), accuracy ($3$\%), gain ($2$\%), length of rule ($2$\%), and the number of attributes per rule ($3$\%). Interestingly, all methods use the number of rules, run time, support, and confidence as evaluation metrics, while only the optimization method employs all metrics (see Figure \ref{metricsdis}). Some papers on discretization methods use the number of frequent itemsets as a metric \cite{Gyenesei2001AFA, Dong2014}. The discretization method primarily uses run time metric with different parameters, such as over the number of records or buckets \cite{lent1997clustering, Miller1997, buchter1998, Kuok1998, fukuda1999mining, Gyenesei2001AFA, lian2005efficient, Guo2008, Yang2010}, minimum support, and minimum confidence \cite{buchter1998,fukuda1999mining, zheng2014optimized, Dong2014, Song2013, brin1999mining} over the number of buckets \cite{buchter1998}, number of sparse points, number of dense regions and number of attributes \cite{lian2005efficient}. On the other hand, the statistical method primarily uses run time \cite{aumann2003statistical, webb2001discovering} and a number of rules \cite{kang2009bipartition, aumann2003statistical} as evaluation metrics. However, other methods also use minimum support, and confidence except for run time and number of rules \cite{ke2008information, hu2022cognitive, QMVMO}. Mean of interest of missing QARs and variance of interest of missing QARs, the maximum interest of missing QARs were also used for performance evaluation in \cite{ke2008information}. Of the $34$ papers ($52$\%) that employ the number of rules as an evaluation metric, $24$ papers ($36$\%) belong to the optimization method. However, $24$ publications ($36$\%) in all have evaluated the NARM algorithms regarding the execution time, among which $15$ ($23$\%) articles are from the discretization method. 

 \subsection{RQ6 Which datasets are used for experiments by NARM methods?}
\begin{figure}
  \centering
  \includegraphics[width=\linewidth]{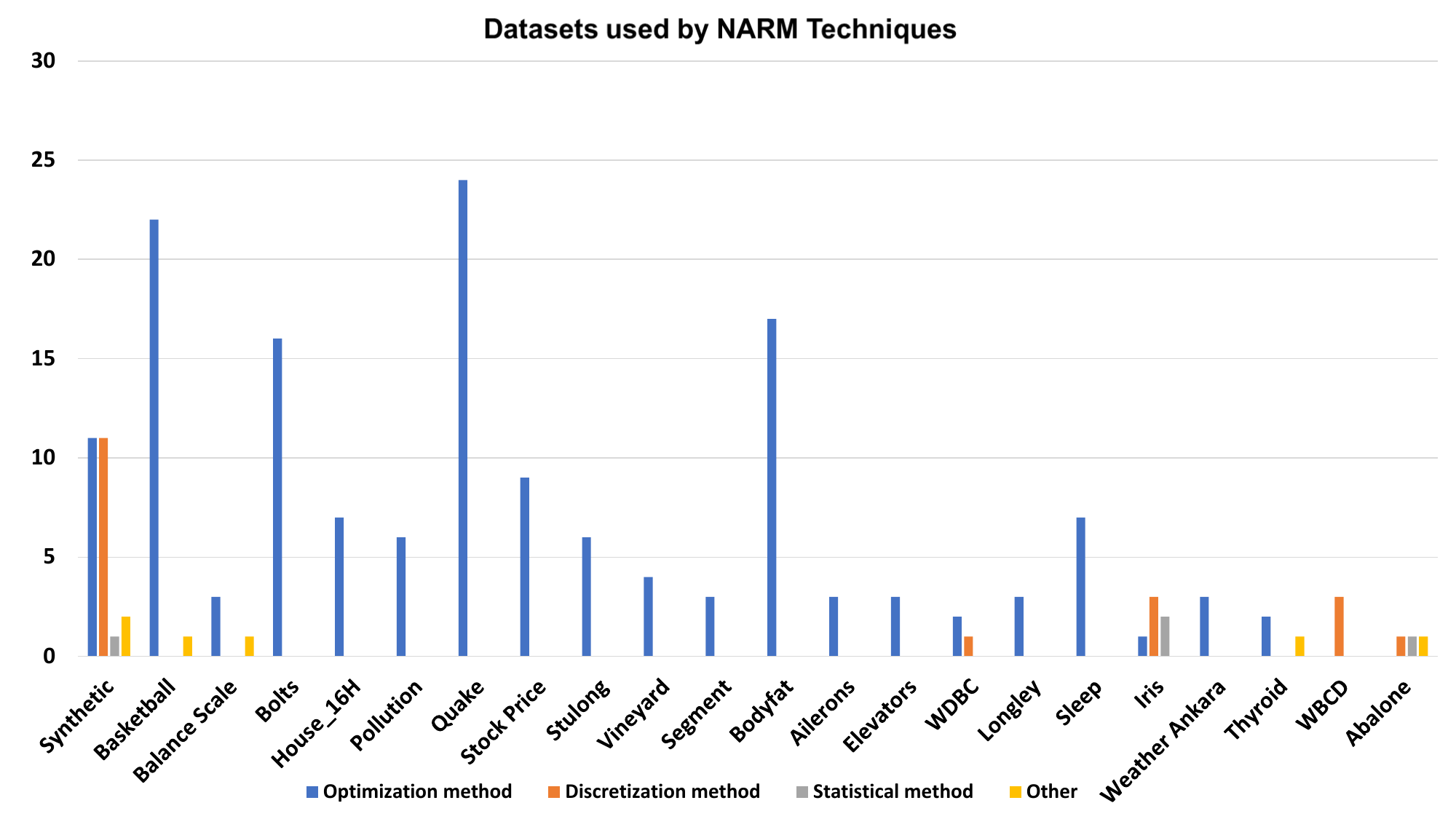}
  \caption{Datasets Used by NARM Techniques.}
  \label{Narmdata}
\end{figure}

Different NARM methods may use different datasets for their experiments, depending on the method type and application domain. Table \ref{dataset} presents the datasets that were most commonly used by different NARM methods, excluding those that were used in only one or two articles. In total, we considered twenty-two datasets, including both real-world and synthetic ones. Fifteen of these datasets were sourced from the Bilkent University Function Approximation Repository (BUFA) \cite{guvenir2000function}, while seven were from the University of California Irvine machine learning repository (UCI) \cite{ucirepo}. Figure \ref{Narmdata} shows the datasets used by NARM methods.

\begin{figure}
  \centering
  \includegraphics[width=\linewidth]{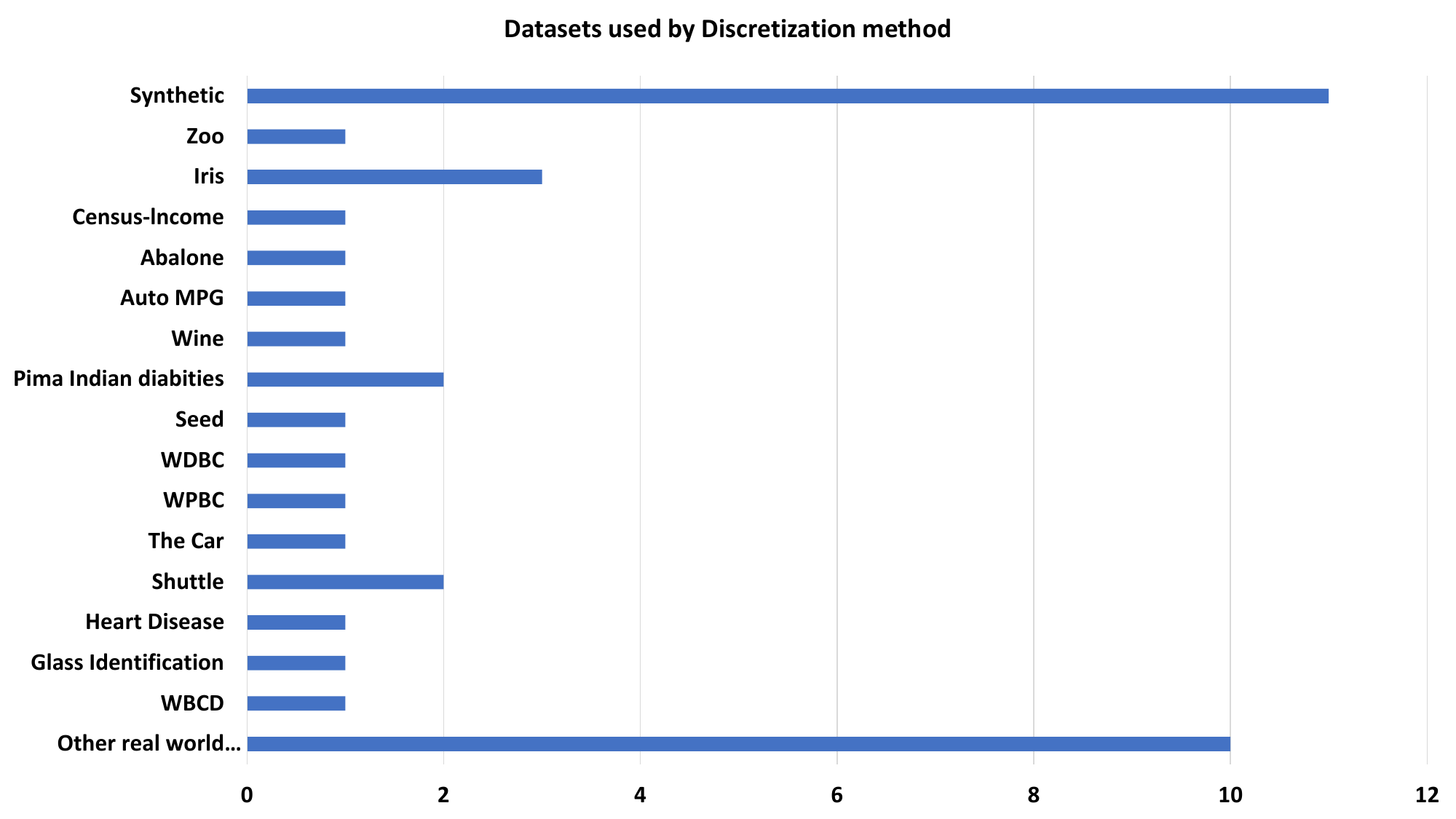}
  \caption{Datasets Used by Discretization Method.}
  \label{discdata}
\end{figure}

The \emph{Quake} dataset was used more frequently than any other dataset, followed by \emph{Basketball}, \emph{Bodyfat}, \emph{Bolts}, and \emph{Stock Price}. Synthetic datasets were also commonly used. Table \ref{datasetd} lists the datasets that were used specifically for the discretization method, which were mostly different from those used by other methods. In total, we considered seventeen datasets, including both real-world and synthetic ones. Most articles on the discretization method used various real-world datasets. As shown in Figure \ref{discdata}, most of these datasets were used in only one article. We also observed that the optimization method articles tended to use datasets from the BUFA repository, while the discretization method articles tended to use datasets from the UCI repository.

\begin{table}
    \centering
    \caption{Data-sets Used by NARM Algorithms (WBC=Wisconsin Breast Cancer Data; WDBC = Wisconsin Diagnostic Breast Cancer) According to Ulitized Methods (Opt.=Optimization; Disc.=Discretization; Stat.=Statistical, Etc.=Other).}
    \label{dataset}
    \begin{tabular} {p{0.8cm}p{2.3cm} p{2.0cm} p{7.4cm}}
    \hline
   Source & Dataset & Methods & References \\ \hline

\multirow{17}{*}{BUFA} & Basketball &  Opt. Etc. & \cite{mata2002discovering,alatacs2006efficient,martinez2011evolutionary,MARTIN2014,martinez2016improving,MARTIN2016,MOEA-QAR,ALATAS2008modenar,ALMASI2015,altay2022chaos,alatas2008rough,ALATASCENPSO,moslehi2011multi,beiranvand2014multi,kahvazadeh2015mocanar,can2017automatic,heraguemi2018mBAT,tahyudin2019improved,kuo2019multi,ledmi2021discrete,moslehi2020novel,QMVMO,Minaeibidgoli2013}\\ 

& Balance Scale & Opt. Etc. & \cite{MARTIN2016,MARTIN2014,ALMASI2015,hu2022cognitive}\\

& Bolts & Opt. & \cite{mata2002discovering,alatacs2006efficient,martinez2011evolutionary,MARTIN2014,MARTIN2014mopnar,martinez2016improving,MARTIN2016,ALTAY2021DEsine,ALATAS2008modenar,ALMASI2015,altay2022chaos,alatas2008rough,moslehi2011multi,tahyudin2019improved,ledmi2021discrete,moslehi2020novel}\\

 & House\_16H & Opt.  & \cite{MARTIN2014,MARTIN2014mopnar,martinez2016improving,MARTIN2016,ALTAY2021DEsine,ALMASI2015,luna2014reducing}  \\

& Pollution & Opt. & \cite{mata2002discovering,alatacs2006efficient,martinez2011evolutionary,MARTIN2014,MARTIN2014mopnar,martinez2016improving,MARTIN2016,ALTAY2021DEsine,ALATAS2008modenar,ALMASI2015,altay2022chaos,alatas2008rough,moslehi2011multi,tahyudin2019improved,moslehi2020novel,ledmi2021discrete}\\

& Quake & Opt. & \cite{mata2002discovering,alatacs2006efficient,martinez2011evolutionary,Minaeibidgoli2013,MARTIN2014,martinez2016improving,MARTIN2016,MOEA-QAR,ALATAS2008modenar,ALMASI2015,altay2022chaos,MARTIN2014mopnar,ALTAY2021DEsine,alatas2008rough,ALATASCENPSO,moslehi2011multi,beiranvand2014multi,kahvazadeh2015mocanar,can2017automatic,heraguemi2018mBAT,tahyudin2019improved,kuo2019multi,ledmi2021discrete,moslehi2020novel}\\

& Stock Price & Opt. & \cite{mata2002discovering,martinez2011evolutionary,MARTIN2014,MARTIN2014mopnar,martinez2016improving,MARTIN2016,ALATAS2008modenar,ALMASI2015,can2017automatic} \\

& Stulong & Opt. & \cite{MARTIN2014,MARTIN2014mopnar,MARTIN2016,ALTAY2021DEsine,ALMASI2015,salleb2007quantminer} \\

& Vineyard & Opt. & \cite{mata2002discovering,martinez2011evolutionary,martinez2016improving,ALTAY2021DEsine}  \\
 
& Segment & Opt. & \cite{MARTIN2014mopnar,MARTIN2016,luna2014reducing} \\

& Bodyfat & Opt. & \cite{mata2002discovering,martinez2011evolutionary,Minaeibidgoli2013,martinez2016improving,MOEA-QAR,ALTAY2021DEsine,ALATAS2008modenar,altay2022chaos,alatas2008rough,ALATASCENPSO,beiranvand2014multi,kahvazadeh2015mocanar,heraguemi2018mBAT,kuo2019multi,tahyudin2019improved,ledmi2021discrete,moslehi2020novel} \\

& Ailerons & Opt. & \cite{ALTAY2021DEsine,martinez2016improving,ledmi2021discrete} \\

& Elevators & Opt. & \cite{ledmi2021discrete,ALTAY2021DEsine,martinez2016improving}\\

& Longley & Opt. & \cite{ALTAY2021DEsine,martinez2016improving,martinez2011evolutionary} \\

& Sleep & Opt. & \cite{mata2002discovering,alatacs2006efficient,martinez2011evolutionary,martinez2016improving,ALATAS2008modenar,alatas2008rough,moslehi2011multi}\\
\hline

\multirow{8}{*}{UCI} & Iris & Opt. Disc. etc. & \cite{salleb2007quantminer,Song2013,DFAC,li,hu2022cognitive,QMVMO} \\

& Weather Ankara & Opt.  & \cite{martinez2016improving,ledmi2021discrete,luna2014reducing} \\
 
& Thyroid & Opt. Etc. & \cite{ke2008information,can2017automatic,MARTIN2016}\\

& WBC & Disc. & \cite{Miller1997,li,DFAC}  \\

& WDBC &  Opt. Etc. & \cite{luna2014reducing,MARTIN2016,zheng2014optimized}    \\

& Abalone &  Disc. Stat. Etc. & \cite{webb2001discovering,Song2013,QMVMO} \\

& Synthetic & Opt. Stat. Etc. & \cite{mata2001mining,mata2002discovering,alatacs2006efficient,salleb2007quantminer,martinez2010mining,alvarez2012evolutionary,ALATAS2008modenar,taboada2008association,srikant1996,lent1997clustering,wang1998interestingness,Kuok1998,zhang1999mining,mg2000,fukuda1999mining,Lee2001,brin1999mining,rastogi2002mining,lian2005efficient,ke2008information,QMVMO,kang2009bipartition,alatas2008rough,ALATASCENPSO,moslehi2011multi} \\

\hline
\end{tabular}
\end{table}
    
\begin{table}
    \centering
    \caption{Data-sets Used by Discretization Method for NARM Algorithms}
    \label{datasetd}
    \begin{tabular} {p{6.5cm}  p{5.5cm}}
    \hline
    Datasets &  References \\ \hline

Wisconsin Breast Cancer Data & \cite{Miller1997,li,DFAC}\\

Wisconsin Prognostic Breast Cancer & \cite{zheng2014optimized}\\

Wisconsin Diagnostic Breast Cancer & \cite{zheng2014optimized}\\

Glass Identification & \cite{li}\\

Heart Disease & \cite{li}\\

Shuttle & \cite{Yang2010,Guo2008}\\

The Car & \cite{Guo2008}\\

Seed & \cite{DFAC}\\

Pima Indian diabities & \cite{zheng2014optimized,Song2013}\\

Wine & \cite{DFAC}\\

Auto MPG & \cite{Song2013}\\

Abalone & \cite{Song2013}\\

Census-lncome & \cite{chan1997effective}\\

Iris & \cite{Song2013,DFAC,li}\\

Zoo & \cite{chan1997effective}\\

Synthetic & \cite{srikant1996,lent1997clustering,wang1998interestingness,Kuok1998,zhang1999mining,mg2000,fukuda1999mining,Lee2001,brin1999mining,rastogi2002mining,lian2005efficient}\\

Other real-world datasets  & \cite{buchter1998,chan1997mining,hong1999mining,mg2000,Gyenesei2001AFA,Mohamadlou2009,Dong2014,Medjadba2019,SWP2019}\\
\hline
\end{tabular}
\end{table}

\subsection{RQ7 What are the challenges and potential future perspectives for the area of NARM?}

To address this research question, a manual identification of the existing research challenges in NARM was conducted. Additionally, the focus was placed on identifying future directions for NARM research.

\subsubsection{Research Challenges} 
After a comprehensive analysis of various NARM methods in both static and dynamic settings, we have identified several issues that NARM needs to address.

\begin{itemize}
    
    \item \emph{Handling Skewed Data:}
     NARM faces challenges when dealing with skewed data, where the data distribution is uneven. Finding associations between numerical variables in such datasets can be difficult and lead to biased results, as well as a high number of irrelevant rules. Moreover, skewed data can have a negative impact on the accuracy and reliability of the analysis, potentially resulting in biased conclusions. Calculation of support and confidence measures can be particularly affected, leading to inaccurate values and erroneous assessments of rule strength. Furthermore, processing skewed data can also impact the speed and efficiency of the algorithms, as they may need to handle a large number of extraneous rules.

    \item \emph{Handling a Large Number of Rules:}
    The main objective of mining numerical association rules is to discover relationships between numerical variables in large datasets. However, this often results in a vast number of association rules, which can make the process computationally expensive, time-consuming, and difficult to sift through to identify the most relevant or interesting rules. To address this challenge, several techniques have been developed to simplify the process and make it more manageable. Some of these techniques include data sampling, the use of efficient algorithms, parallel and distributed computing, dimensionality reduction, and pruning methods. By implementing these techniques, it becomes easier to extract useful association rules from large datasets, reduce the size of the dataset, simplify the mining process, and speed up computations.

    \item \emph{Quality of Association Rules:}
    Extracting high-quality rules is also a challenge in NARM due to the potential for redundancy, irrelevance, and conflicts in the rules. The large and complex nature of the datasets used in NARM can lead to a large number of rules, making it difficult to identify the most relevant ones. Additionally, skewed data can impact the reliability of support and confidence measures, further affecting the quality of the rules. The rules generated by NARM algorithms may also be challenging to interpret and understand. To address these issues, data pre-processing, the use of alternative metrics, the selection of appropriate algorithms, and the application of ensemble methods can be helpful in improving the quality of association rules.
    
    \item \emph{Complex Relationship:}
    Numerical data often contain intricate relationships, such as non-linear or multi-dimensional relationships, which can be difficult to represent and analyze using traditional ARM algorithms. This may result in inaccurate or incomplete rules, which can impact the reliability and accuracy of the analysis. To address this challenge, advanced algorithms such as decision trees, artificial neural networks, or support vector machines can be utilized in NARM. Ensemble approaches like gradient boosting or random forests can also be helpful in addressing complex relationships by combining the output of multiple algorithms to produce more accurate results. However, these techniques may increase computational complexity and require more data and computing resources to be effective.

    \item \emph{Handling Outliers:}
    Outliers are extreme values that differ significantly from the majority of values in the dataset and can impact the accuracy and reliability of the results of ARM. Outliers may indicate genuine data variances, or they could result from measurement errors or data input issues. Several methods can address this problem, including outlier detection, data cleaning, data transformation, and robust algorithms. These methods can help remove or mitigate the effect of outliers, ensuring that the mining process yields more accurate and reliable results.

\end{itemize}

\subsubsection{Future directions}

\begin{itemize}

\item \emph{Handling Big Data:} 
Despite conducting a thorough SLR, we were unable to identify any studies that focus on retrieving numerical association rules from big data. However, the rise of big data will undoubtedly have a significant impact on the future of NARM. Developing more efficient algorithms that can handle vast amounts of numerical data will be essential as big data continues to become increasingly common. This will likely lead to the development of new algorithms specifically designed for big data that are optimized for scalability, speed, and  accuracy. Additionally, advanced data cleanings and preprocessing techniques, such as outlier detection, imputation of missing values, and feature selection, will become increasingly important to ensure the quality of results.

    \item \emph{Explainable AI:}
Improving the interpretability and explainability of NARM results is a critical research direction. Explainable AI \cite{BARREDOARRIETA2020, fister2023comprehensive} can enhance transparency and comprehension, which is essential for non-experts to validate the findings and ensure their alignment with the intended objectives. By revealing the underlying reasoning behind the results, Explainable AI can assist users in making better decisions and recognizing any inherent biases or limitations in the outcomes. Therefore, developing NARM techniques that provide transparent and comprehensible results is a vital area of research.

    \item \emph{Hybrid Approach:}
A promising future direction in NARM is to leverage the strengths of various methods and techniques through hybrid approaches. Some studies, such as \cite{moslehi2020novel, altay2022chaos, ALTAY2021DEsine,zhang1999mining, Mohamadlou2009, Wang2015, kianmehr2010fuzzy}, have attempted to combine different approaches to improve the results of NARM. Combining NARM with deep learning, integrating rule-based and distance-based approaches, and combining unsupervised and supervised learning are all hybrid approaches that show potential in this field. By integrating these techniques, the limitations of individual methods can be addressed, resulting in a more accurate and thorough analysis of the relationships between variables in numerical data. Hybrid approaches can lead to valuable insights and more reliable results.

    \item \emph{Handling Streaming Data:}
To keep up with the increasing demand for real-time analysis, developing NARM algorithms that can handle streaming data and update association rules in near real-time is crucial. In applications where timely and accurate decisions are critical, streaming data enables real-time analysis of numerical data, which allows organizations to make informed decisions based on up-to-date information. The ability to analyze a larger volume of numerical data in real time will lead to more comprehensive and accurate results. Moreover, streaming data enables dynamic updates to the results of NARM as new data becomes available, providing a more accurate and comprehensive view over time. Therefore, developing NARM algorithms that can handle streaming data is an important future direction.

    \item \emph{Incorporating Machine Learning Techniques:}   
      The integration of machine learning techniques, such as deep learning, into NARM, has the potential to revolutionize the field. With the ability to automatically detect patterns and relationships in the data, which may not be immediately apparent to human analysts, machine learning algorithms can significantly enhance the accuracy of the results.  Moreover, this approach can reduce the time and effort required to identify such patterns in the data. The utilization of machine learning can also expand the scope of NARM applications across various industries, as these algorithms can handle complex data more efficiently, including high-dimensional data or data with non-linear relationships.

    \item \emph{Privacy and Security:}
The importance of privacy and security in NARM is increasing, and it is imperative to protect and use data ethically. However, the existing studies in this SLR did not address these issues. To ensure data protection, sensitive information can be removed or masked using anonymization techniques while preserving the necessary data for ARM. Furthermore, to reduce the risk of unauthorized access, the data can be partitioned into smaller subsets, and access control methods can be developed to control who has access to the data and the association rules generated from it. Incorporating these privacy and security measures will safeguard the data and ensure its ethical use.

  \end{itemize}  
    
 \subsection{RQ8 How to automate discretization of numerical attributes for NARM in a useful (natural) manner?}

 
Developing novel methods and techniques for NARM is a continuous area of research, and the discretization method serves as the foundation for NARM \cite{srikant1996}. However, selecting the best partitions for discretizing complex real-world datasets still lacks a benchmark method. None of the discretization methods that we figured out with the review, see Sects. \ref{TheDiscretizationMethod} in \ref{TheDiscretizationMethodALGOS}, explicitly addresses human perception of partitions. Therefore, we proposed a novel discretization technique in our research \cite{impact}, which utilizes two measures for order-preserving partitioning of numerical factors: the \emph{Least Squared Ordinate-Directed Impact Measure} (LSQM) and the \emph{Least Absolute-Difference Ordinate-Directed Impact Measure} (LADM).

These proposed measures offer a straightforward method for finding partitions of numerical attributes that reflect best the impact of one independent numerical attribute on a dependent numerical attribute. We thoroughly experimented with these measures and compared the outcomes with human perceptions of partitioning a numerical attribute \cite{discretizing}. To develop an automated measure for discretizing numerical attributes, understanding perceptual conception is crucial. To achieve this, we investigated the impact of data points' features on human perception when partitioning numerical attributes \cite{Analysis}. These efforts have contributed to the development of more accurate and efficient methods for NARM.

\section{Threats to validity}
\label{threats}
This section outlines potential threats to the validity of this SLR that might bias the outcomes of our in-depth investigation. The first threat pertains to defining the search string. We make no claims regarding the perfection of the search string used in the process. While we included all relevant search terms related to NARM, it is possible that the search terms may not have captured all relevant NARM-related work. To mitigate this risk, we included synonyms for ``numerical association rule mining'' and abbreviations such as ``NARM'' and ``QARM'' in the search term.
The second threat pertains to the selection of digital libraries to search for articles. Although we searched five digital libraries for computer science, it is possible that additional sources may have produced different outcomes. To minimize bias, we manually searched Google Scholar and also looked through the list of references for the selected primary studies to identify significant publications. We are confident that the majority of published research on NARM is covered in this study. The third threat is the inclusion and exclusion of articles. To determine whether a paper should be included or excluded, we first reviewed the title, abstract, and keywords according to the inclusion and exclusion criteria. Then, we manually checked for references to ensure we did not miss any relevant papers. Additionally, we evaluated the selected studies using a quality assessment procedure. The fourth threat is about the time frame. Since the search process was conducted in early 2022, only articles published between 1996 and the beginning of 2022 were included. It is possible that we may have missed some articles published after the specified time frame.
The final threat concerns article selection. The authors of this article may have been biased in their choice and categorization of publications that were included. Two authors chose articles based on their personal experiences. Although the final selection was made by a single author, all studies were verified by other authors to minimize bias.

\section{Discussion and Finding}
\label{discussion}

\begin{figure}
  \centering
  \includegraphics[width=\linewidth]{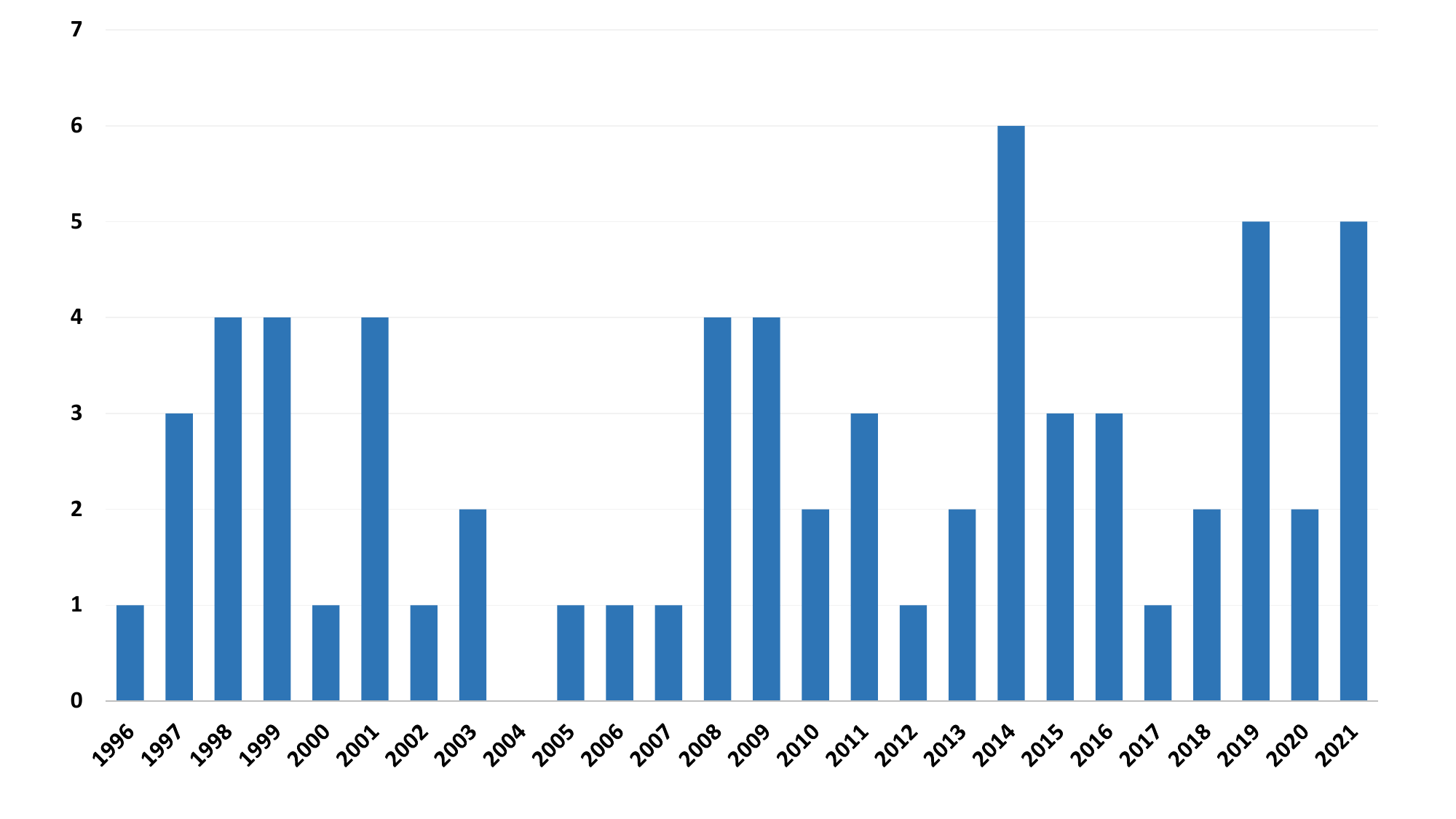}
  \caption{Distribution of Articles by Publication Year.}
  \label{distyear}
\end{figure}

In this SLR, we analyzed a total of $68$ studies on NARM published between 1996 and 2022. Our analysis revealed several significant findings and trends. Figure \ref{distyear} illustrates the distribution of selected articles by publication year, with a notable concentration in 2014 and strong NARM trends in 2019 and 2021.

\begin{figure}
\centering
\begin{subfigure}{.58\textwidth}
  \centering
  \includegraphics[width=1\linewidth]{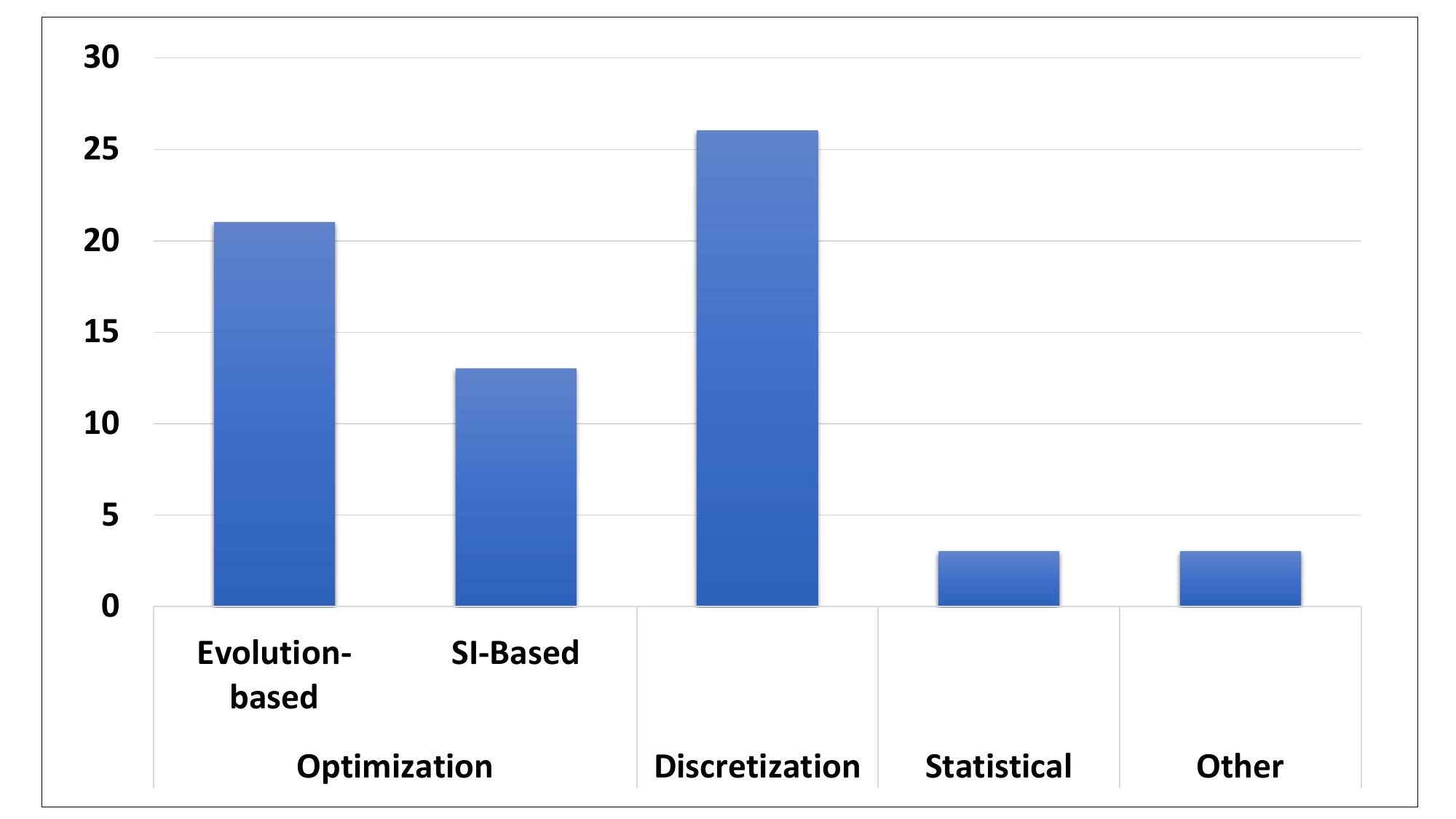}
  \caption{Distribution of the Articles by Method.}
  \label{fig:sub1}
\end{subfigure}%
\begin{subfigure}{.58\textwidth}
  \centering
  \includegraphics[width=1\linewidth]{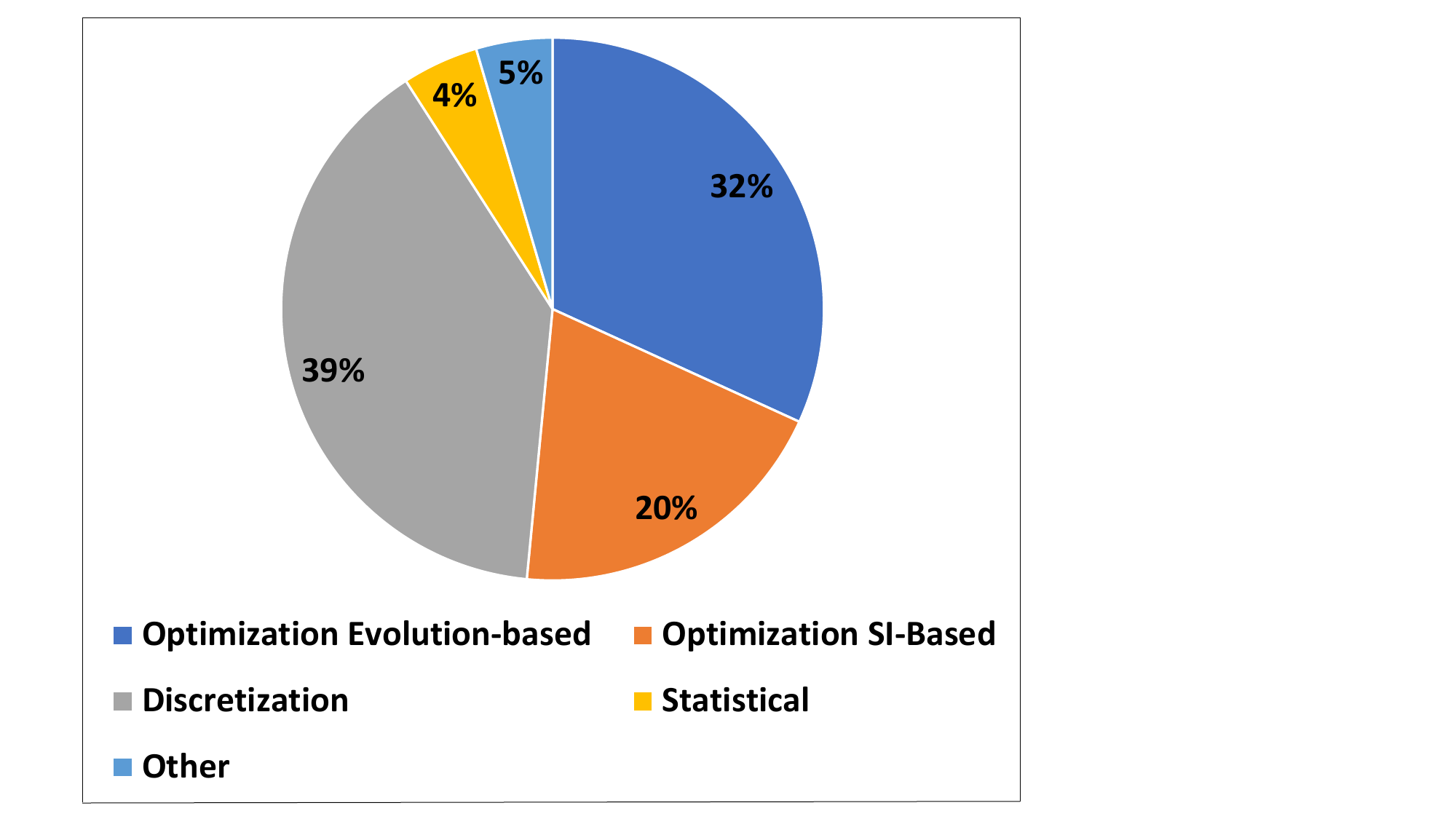}
  \caption{Proportions of the Articles.}
  \label{fig:sub2}
\end{subfigure}
\caption{Distribution of the Primary Study.}
\label{distribution}
\end{figure}

\begin{figure}
\centering
  \begin{subfigure}{.58\textwidth}
  \centering
  \includegraphics[width=1\linewidth]{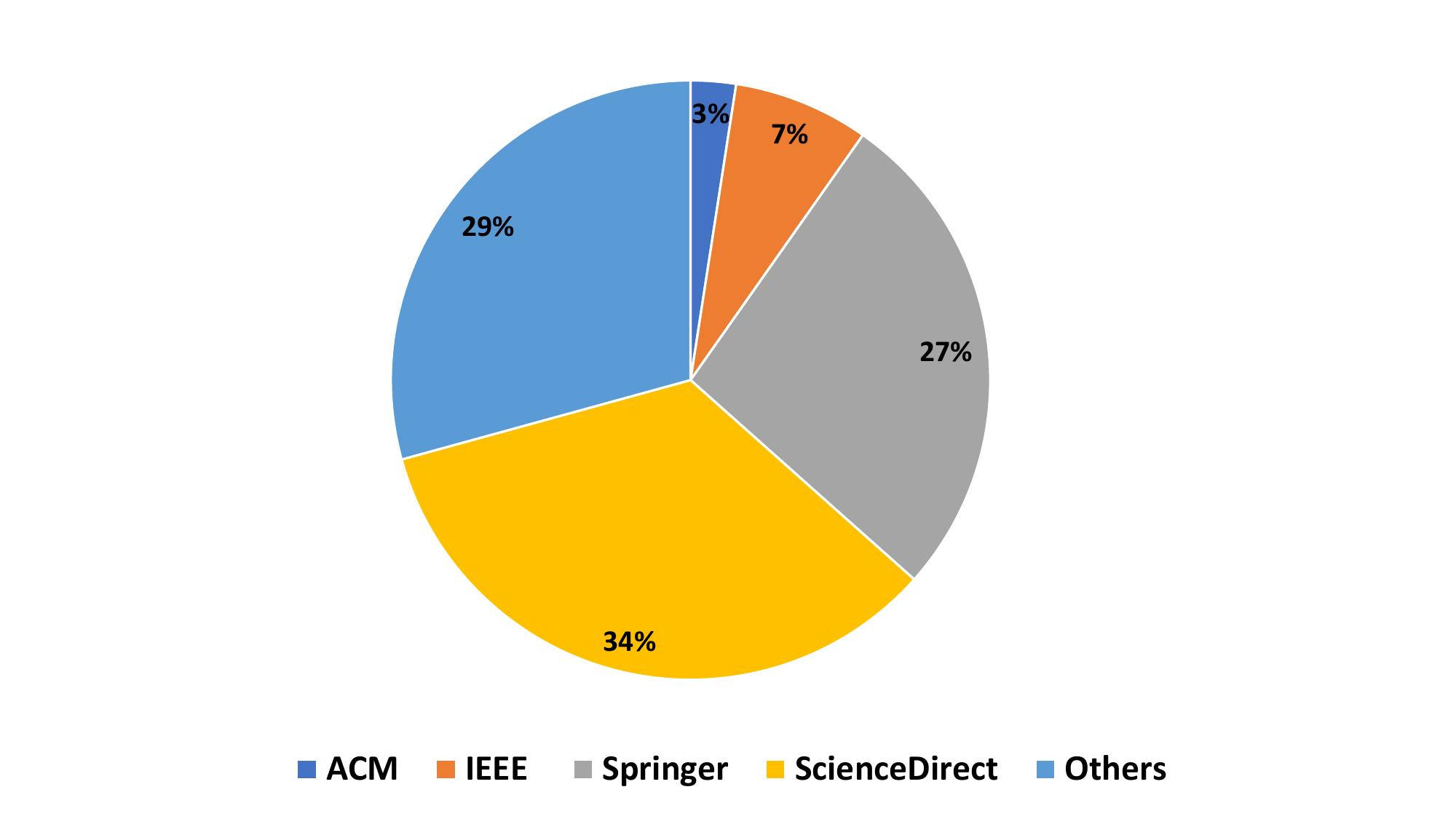}
  \caption{Articles Published in Journals.}
  \label{fig:sub3}
  \end{subfigure}%
  \begin{subfigure}{.58\textwidth}
  \centering
  \includegraphics[width=1\linewidth]{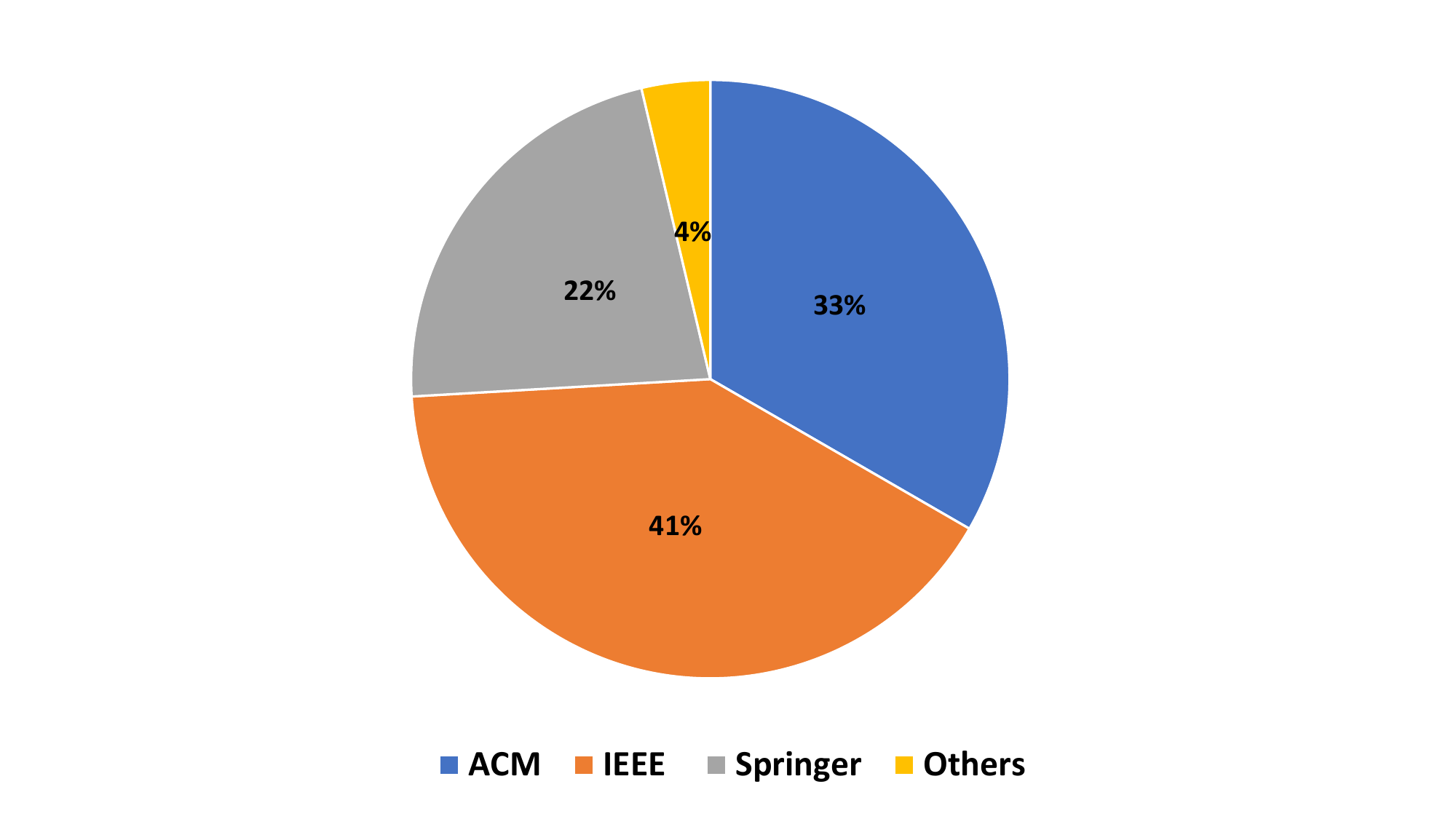}
  \caption{Articles Published in Conferences.}
  \label{fig:sub4}
  \end{subfigure}
  \begin{subfigure}{.58\textwidth}
  \centering
  \includegraphics[width=1\linewidth]{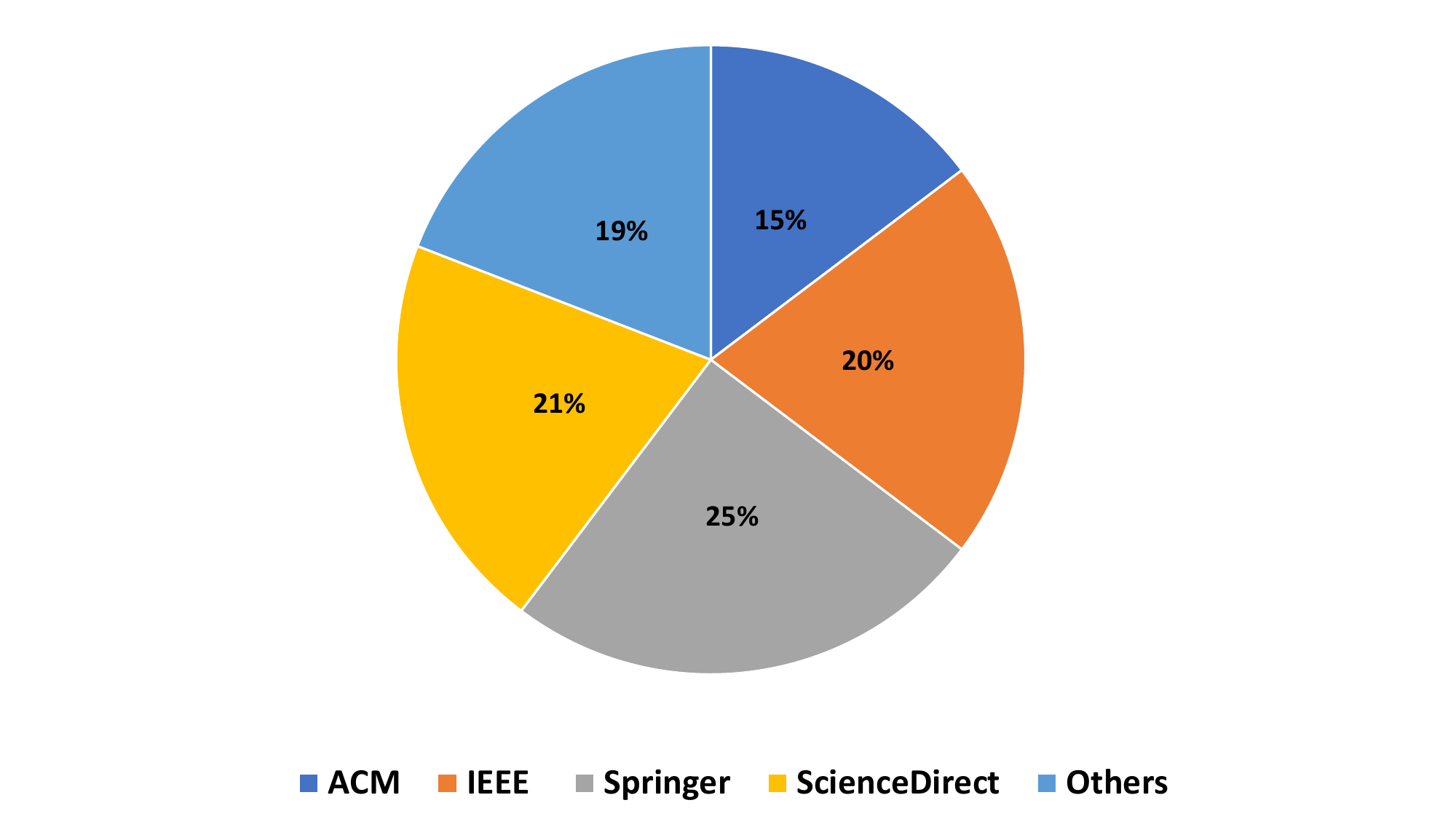}
  \caption{Total Articles Published in Different Sources.}
  \label{fig:sub5}
  \end{subfigure}%
\caption{Publication Source Distribution: 1996-2022.}
\label{publishedarticle}
\end{figure}

\begin{table}
    \centering
    \caption{Type of Publications in the Area of NARM}
    \label{publication}
    \begin{tabular} {p{4cm} p{4cm} p{3cm}}
    \hline
    Publication Source &  Type of Publication & Number of Articles\\ \hline
IEEE  & Journal & 3 \\
Springer & Journal & 11 \\
ACM & Journal & 1\\
ScienceDirect  & Journal & 14 \\
Other & Journal & 12 \\

IEEE  & Conference & 11 \\
Springer & Conference & 6 \\
ACM & Conference & 9 \\
Other & Conference & 1 \\
\hline
 \end{tabular}
\end{table}

\begin{table}
    \centering
    \caption{Advantages and Limitations of NARM Methods}
    \label{advantage}
    \begin{tabular} {p{2.5cm} p{5cm} p{5cm}}
    \hline
    NARM Method &Advantages & Limitations\\ \hline
    The Discretization method & Easier to interpret. Efficient and scalable to process
    huge datasets. Able to deal with continuous variables. Highly flexible. &  Requirement of a user-specified threshold. Information loss. Unable to deal with high dimensional data. Discretization bias can lead to inaccurate or unreliable results.
    Membership functions should be known in advance.\\
    The Optimization method &  No need to determine the minimum support and minimum confidence. No need for a prior discretization step. High scalability.  & Higher computational cost. Low search capability in the local area. Stuck in local minima. Convergence issues.\\
 
    The Statistical method & Measuring significance to identify meaningful association rules. Can handle noise and missing data. Provide quantifiable results. & Lack of scalability. Not able to detect complex relationships in data. Not suitable for handling high-dimensional data. Not well-suited for mining rules in numerical data.          \\
     \hline
\end{tabular}
\end{table} 

 \begin{figure}
  \centering
  \includegraphics[width=\linewidth]{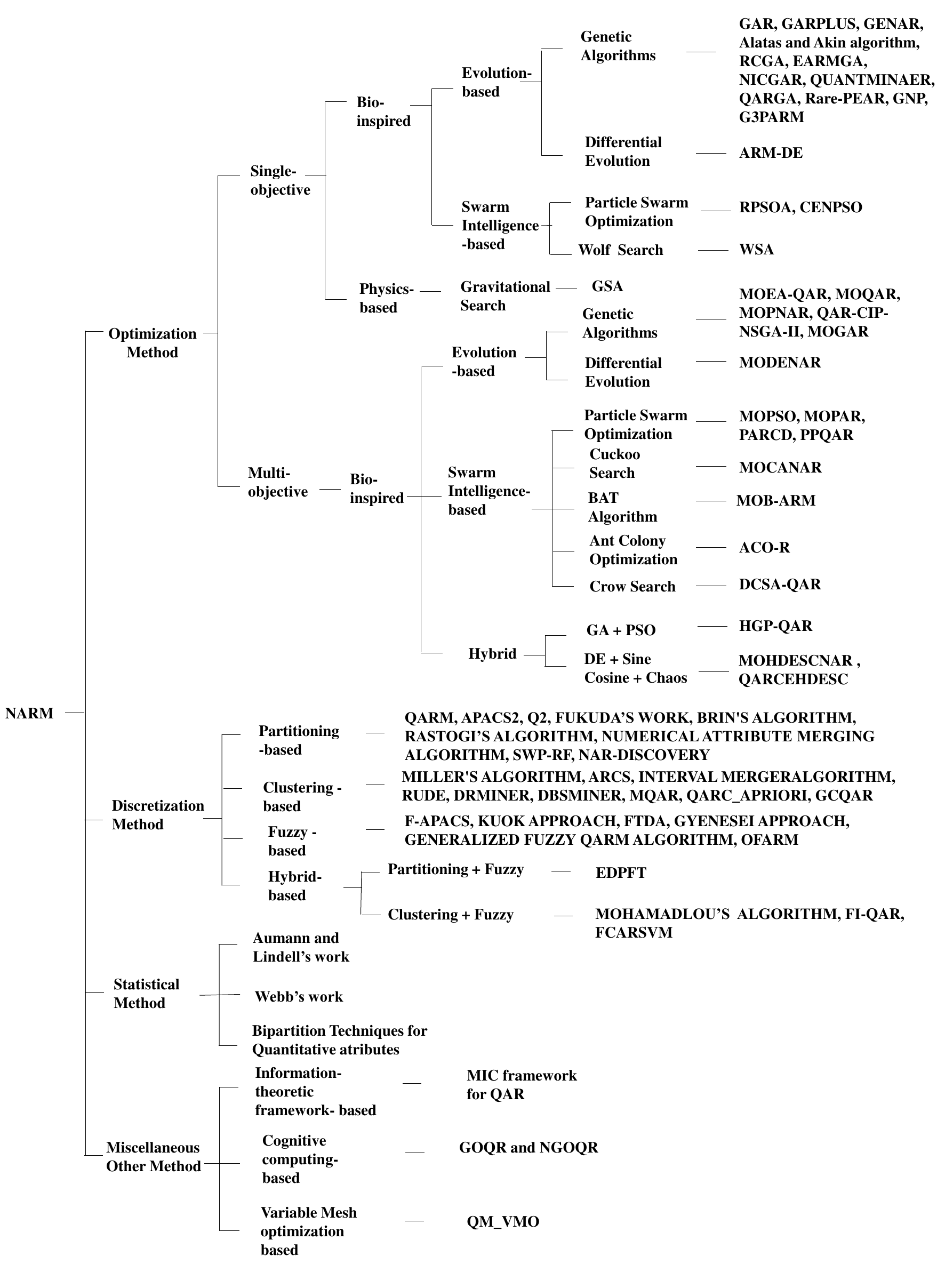}
 \caption{Visual Exploration and Illustration of NARM Methods and Algorithms: A Comprehensive Overview.}
  \label{NARM}
\end{figure}

The distribution of primary studies by the method is presented in Figure \ref{distribution}. The majority of articles focus on the discretization method, followed by the evolution-based optimization method (Figure \ref{fig:sub1}). Statistical and other techniques make up a smaller portion. We further detail the partition of articles by the method as a percentage in Figure \ref{fig:sub2}. Table \ref{publication} provides the number of articles published in different journals and conferences. Overall, $60\%$ of the selected articles were published in journals and $40\%$ in conferences. Figure \ref{publishedarticle} illustrates the percentage of articles published in journals and conferences.
ScienceDirect published $34\%$ journal articles (Figure  \ref{fig:sub3}) however, IEEE published $41\%$ conference articles (Figure \ref{fig:sub4}), which is the highest number and Springer published $25\%$ overall articles (Figure \ref{fig:sub5}).
We identified several methods for solving NARM, including the discretization method, optimization methods using evolutionary and bio-inspired approaches, the statistical method, and other methods. The discretization method was the most widely used, accounting for $39\%$ of the total articles (Figure \ref{fig:sub2}).
Evolution-based and SI-based optimization methods covered $32\%$ and $20\%$ of the total articles, respectively (Figure \ref{fig:sub2}). The optimization and discretization methods had a greater impact compared to the statistical and other methods. Each method also employs various approaches to address NARM problems. For example, the discretization method includes the clustering approach, which encompasses density-based and grid-based techniques. 
The partition approach, which involves converting continuous numerical data into discrete values by grouping them into intervals or bins, was found to be simple and easy to implement. However, the choice of interval or bin size can affect result accuracy, and it may not be suitable for datasets with a large number of variables.  The optimization method encompasses a range of approaches, such as genetic algorithms, grammar-guided genetic programming, differential evolution, particle swarm optimization, gravity-based algorithms, swarm-based algorithms, Cauchy distribution, and hybrid-based methods. The statistical method is limited and utilizes various distribution scales, including mean, median, variance, and standard deviation. Some studies did not fit into any specific method and were categorized as miscellaneous other methods  based on the information-theoretic approach, cognitive computing, and variable mesh.

Many algorithms \cite{lent1997clustering,mg2000,hong1999mining,zhang1999mining,zheng2014optimized,srikant1996,Wang2015} based on the discretization method use apriori algorithm for generating  association rules.  However, the evolution and SI-based algorithms do not use the apriori algorithm. Additionally, certain algorithms under the discretization method have employed new measures, including density measure \cite{lian2005efficient,Guo2008}, R-measure \cite{zhang1999mining}, Certainty Factor \cite{zheng2014optimized}, and adjusted difference measure \cite{chan1997effective}. Figure \ref{NARM} demonstrates the visual presentation of NARM methods and their algorithms.
 In Table \ref{advantage}, we have summarized the advantages and limitations of each method. 

 Our analysis found that most studies based on the discretization method used synthetic and real-world data to evaluate the effectiveness of NARM algorithms. However, evolutionary and SI-based algorithms mostly used common datasets, such as \emph{Quake}, \emph{Basketball}, \emph{Bolt}, \emph{Bodyfat}. Furthermore, the \emph{Iris} dataset was the only one commonly used by discretization, optimization, and statistical methods-based algorithms. It is crucial to note that the choice of the dataset may impact the performance of the methods, and further studies are needed to evaluate the algorithms' practical applicability on real-world datasets.

In our extensive review of the literature on NARM, we analyzed various metrics used to evaluate the performance and effectiveness of different algorithms and models. Our study focused on important metrics such as generated number of rules, run time, and value of support and confidence. Support measures the frequency of a specific item set or rules in the dataset, frequently used in conjunction with confidence, which quantifies how likely a certain outcome is given an antecedent.
However, it is crucial to carefully interpret and evaluate the reliability and validity of the metrics used, as they can lead to spurious or irrelevant associations if not used properly.
Our SLR sheds light on the most commonly used metrics in NARM, including those used by multi-objective algorithms.

Multi-objective NARM algorithms consider different objectives simultaneously to generate a set of Pareto-optimal solutions that balance competing objectives. The choice of objective
for multi-objective NARM algorithms depends on the research question and data characteristics. Some common objectives include maximizing support or confidence while minimizing the number of rules generated.

As a rapidly evolving research area, NARM presents numerous potential future directions for research. These include exploring new scalable optimization algorithms, addressing Big Data challenges, incorporating explainable AI into the mining process, integrating machine learning techniques, addressing security concerns, and using hybrid approaches. By pursuing these directions, researchers can advance the state of the art in NARM and develop more effective and practical solutions for real-world applications.

\section{Conclusion}
\label{conclusion}
This article addressed a significant research gap in the field of NARM and provided readers with a comprehensive understanding of the state-of-the-art methodologies and developments in the domain. Moreover, this study serves as a foundation for future research and offers comprehensive insights for researchers working on NARM-related problems. To achieve this, a comprehensive SLR is conducted based on the guidelines set forth by Kitchenham and Charter. We conducted a detailed examination of a wide range of methods, algorithms, metrics, and datasets sourced from 1,140 scholarly articles spanning the period from the introduction of NARM in 1996 to 2022. Eventually, through a rigorous selection process, including several inclusion, exclusion and quality assessment criteria, 68 articles were selected for this SLR. By providing an exhaustive understanding of the existing NARM methods, highlighting their strengths and limitations, as well as identifying research challenges and future directions, we aim to stimulate innovative thinking and encourage the exploration of novel approaches in NARM. These perspectives include exploring new scalable optimization algorithms, analyzing NARM methods with big data, incorporating explainable AI into the mining process, incorporating machine learning techniques, addressing security concerns, and using hybrid approaches. Subsequently, based on the finding of this SLR, a novel discretization measure is presented to aid in NARM that explicitly addresses the human perception of partitions. The ultimate goal of this review is to inspire and guide researchers in developing more effective and practical solutions for real-world NARM applications.

\begin{acks}
This work has been partially conducted in the project ``ICT programme'' which was supported by the European Union through the European Social Fund.
\end{acks}
\section*{Declarations}
\textbf{Conflict of interest} Authors declare that they have no conflict of interest.
%
\bibliographystyle{ACM-Reference-Format}
\bibliography{NARM-SLR}

%
\appendix

\end{document}